\documentclass[runningheads]{llncs}

\usepackage{eccv}

\usepackage{eccvabbrv}

\usepackage{graphicx}
\usepackage{booktabs}
\usepackage{svg}

\usepackage[accsupp]{axessibility}  % Improves PDF readability for those with disabilities.

\usepackage{hyperref}

\usepackage{multirow}
\usepackage{colortbl}
\usepackage{soul}
\usepackage{makecell}
\usepackage{orcidlink}

\definecolor{customgreen}{rgb}{0.25, 0.69, 0.65}
\definecolor{customyellow}{rgb}{1, 1, 0.38}

\begin{document}

% ---------------------------------------------------------------
% TODO REVIEW: Replace with your title
\title{Are Synthetic Data Useful for Egocentric Hand-Object Interaction Detection?} 

% TODO REVIEW: If the paper title is too long for the running head, you can set
% an abbreviated paper title here. If not, comment out.
\titlerunning{Are Synthetic Data Useful for Egocentric HOI Detection?}

% TODO FINAL: Replace with your author list. 
% Include the authors' OCRID for the camera-ready version, if at all possible.
\author{Rosario Leonardi\inst{1}\orcidlink{0009-0001-8693-3826} \and Antonino Furnari\inst{1,2}\orcidlink{0000-0001-6911-0302} \and \\ Francesco Ragusa\inst{1,2}\orcidlink{0000-0002-6368-1910} \and Giovanni Maria Farinella\inst{1,2}\orcidlink{0000-0002-6034-0432}}

% TODO FINAL: Replace with an abbreviated list of authors.
\authorrunning{R.~Leonardi et al.}
% First names are abbreviated in the running head.
% If there are more than two authors, 'et al.' is used.

% TODO FINAL: Replace with your institution list.
\institute{Department of Mathematics and Computer Science, University of Catania, Italy \and Next Vision s.r.l., Italy}

\maketitle

\begin{abstract}
In this study, we investigate the effectiveness of synthetic data in enhancing egocentric hand-object interaction detection. 
Via extensive experiments and comparative analyses on three egocentric datasets, \textit{VISOR}, \textit{EgoHOS}, and \textit{ENIGMA-51}, our findings reveal how to exploit synthetic data for the HOI detection task when real labeled data are scarce or unavailable. 
Specifically, by leveraging only $10\%$ of real labeled data, we achieve improvements in \textit{Overall AP} compared to baselines trained exclusively on real data of: $+5.67\%$ on \textit{EPIC-KITCHENS VISOR}, $+8.24\%$ on \textit{EgoHOS}, and $+11.69\%$ on \textit{ENIGMA-51}.
Our analysis is supported by a novel data generation pipeline and the newly introduced \textit{HOI-Synth} benchmark which augments existing datasets with synthetic images of hand-object interactions automatically labeled with hand-object contact states, bounding boxes, and pixel-wise segmentation masks. Data, code, and data generation tools to support future research are released at: \url{https://fpv-iplab.github.io/HOI-Synth/}.
\keywords{Synthetic Data \and Egocentric HOI \and Domain Adaptation}
\end{abstract}

% ----------------------
% ---- INTRODUCTION ----
% ----------------------

\section{Introduction}
\label{sec:intro}

\begin{figure}[t]
    \centering
    \includegraphics[width=\linewidth]{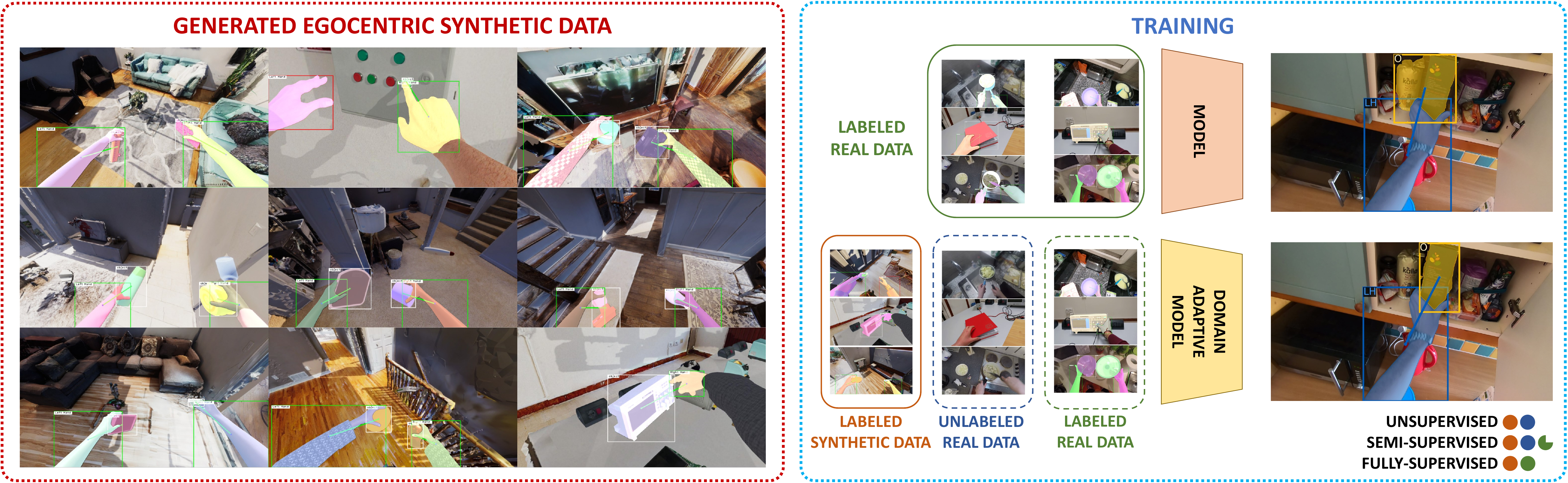}
    \caption{We study the impact of synthetic data in egocentric hand-object interaction detection. We generate and automatically label large sets of synthetic data (left) and study a set of domain adaptation scenarios in which models are trained on both synthetic and real unlabeled data, plus different amounts of labeled real data (right).}
    \label{fig:figure1}
\end{figure}

Understanding how humans interact with the surrounding objects from egocentric images is a fundamental challenge in computer vision, with applications in diverse domains including collaborative robotics~\cite{edsinger2007human, carfi2021hand}, industrial behavior understanding~\cite{sener2022assembly101, Ragusa2021TheMD}, human-computer interaction~\cite{lv2022deep}, and healthcare~\cite{besari2023hand}.
Previous works investigated the task of understanding human-object interactions from an egocentric perspective in different forms, including action recognition~\cite{Damen2021RESCALING}, object state-change detection~\cite{Grauman2021Ego4DAT}, and hand-object interaction forecasting~\cite{liu2022joint}.
A line of work developed around the goal of identifying the actively manipulated object, the presence of hands, and the contact state between hands and objects~\cite{Shan2020UnderstandingHH, Ragusa2021TheMD, leonardi2022egocentric, VISOR2022}, which is generally referred to as \textit{Hand-Object Interaction (HOI) detection}.
Despite the progress in model design granted by the availability of egocentric benchmarks such as EPIC-KITCHENS~\cite{damen2018scaling, Damen2021RESCALING} and VISOR~\cite{VISOR2022}, performance in real application scenarios is closely tied to the availability of large amounts of annotated real-world and domain-specific data~\cite{Ragusa2021TheMD}. In addition, the need for spatial and interaction annotations makes acquiring and labeling such data an expensive and time-consuming process.

The use of synthetic data to reduce the dependence of prediction algorithms on labeled real data has been previously explored in different domains, including embodied AI~\cite{kolve2017ai2,savva2019habitat,xia2020interactive} and autonomous driving~\cite{Dosovitskiy17,fabbri21iccv}. However, the exploitation of synthetic data is currently under-explored in egocentric vision in general and hand-object interaction detection in particular, due to the challenges associated to generating accurate and photorealistic images of hand-object interactions, which requires the modeling of hands, objects and physical contact.
As a result, many questions still remain unanswered: \textit{1) Is there a gap between real and synthetic data?  2) Where does it originate?  3) How can it be reduced? 4) Can synthetic data entirely replace real data? 5) Can synthetic data enable training in the presence of unlabeled real data?  6) Can synthetic data increase efficiency when few real data are labeled?  7) What scale of synthetic data is needed? 8) Is in-domain synthetic data, aligned to the target real domain in terms of objects and environment, beneficial? }

With the goal of advancing research in egocentric hand-object interaction detection and synthetic-to-real domain adaptation for egocentric vision, in this paper, we propose a systematic investigation to answer the questions above.
To support our investigation, we propose a novel pipeline and develop a simulator able to generate synthetic images of realistic hand-object interactions in multiple environments, which are automatically labeled for the considered hand-object detection task (Figure~\ref{fig:figure1}-left). We generate three sets of synthetic data, paired with two popular domain-generic hand-object detection benchmarks, \textit{EPIC-KITCHENS VISOR}~\cite{VISOR2022}, and  \textit{EgoHOS}~\cite{EgoHos_jianbo_eccv22}, and a domain-specific dataset, \textit{ENIGMA-51}~\cite{ragusa2024enigma}. 
We hence study three different domain adaptation tasks: \textit{unsupervised domain adaptation}, where models are trained with synthetic data and unlabeled real data, \textit{semi-supervised domain adaptation}, where models are trained with synthetic data, unlabeled real data, and few labeled real data, and \textit{fully supervised domain adaptation}, where models are trained with labeled synthetic and real data (Figure~\ref{fig:figure1}-right).
Collectively, the real and generated egocentric data define a new benchmark dataset, which we term \textit{HOI-Synth}. 

We leverage \textit{HOI-Synth} to benchmark different approaches to domain adaptation for hand-object interaction detection based on previous literature on domain adaptation for object detection~\cite{li2022cross,tarvainen2017mean,liu2021unbiased,ganin2015unsupervised} and hand-object interaction detection~\cite{VISOR2022} in multiple settings. Our analysis provides several insights into the advantages of using properly generated synthetic data for egocentric hand-object interaction detection: A) Despite the progress in realistic data generation, a domain gap between synthetic and real data still exists, with models trained only on synthetic data lagging behind models trained on real data by large margins (a gap of $\sim 30\%-40\%$ in AP), which we attribute to limits in photorealism, accuracy of grasping, and diversity of environments and objects; B) We show that performing domain adaptation allows to reduce the synth-real gap in the settings of \textit{unsupervised domain adaptation}, where methods obtain large improvements of $\sim 20\% - 35\%$ AP when exposed to unlabeled real data, \textit{semi-supervised domain adaptation}, where models achieve performance comparable to approaches trained only on real data by using only $\sim 10\%-25\%$ of real data labels, and \textit{fully-supervised domain adaptation}, where combining labeled real and synthetic data improves performance by $\sim 1\% - 4\%$  AP; C) While most of the improvements comes from synthetic sets in the order of $10,000$ images, methods still obtain benefits as the amount of synthetic data is increased up to $30,000$; D) When available, in-domain synthetic data including objects and environments aligned to those of the target real domain, greatly improves performance in the unsupervised domain adaptation setting, with gains of up to $\sim +20\%$ AP, while advantages of in-domain synthetic data are limited if few real labeled data are available for semi-supervised adaptation.

The contributions of this work are: 1) A systematic investigation of the egocentric hand-object interaction detection task assessing the effectiveness of properly generated synthetic data in three domain adaptation settings. Our investigation provides insights into the usefulness of synthetic data and will inform future model and experimental designs; 2) \textit{HOI-Synth}, a novel benchmark for unsupervised,  semi-supervised and fully-supervised domain adaptation which, for the first time, enables the study of synthetic-to-real domain adaptation for egocentric hand-object interaction detection. With the benchmark, we include several baseline results showcasing the potential of synthetic data in this domain and providing a basis for future comparisons and advances; 3) A novel data generation pipeline and a developed simulator, which will be able to support future investigations in the exploitation of synthetic data for egocentric vision. 
To enable future research on this topic, we publicly release the generated data, the simulator, and all the code required to reproduce the results.

% ----------------------
% ---- RELATED WORK ----
% ----------------------

\section{Related Work}
\label{sec:related_work}

\noindent\textbf{Hand-Object Interaction Detection} 
The authors of~\cite{Shan2020UnderstandingHH} were among the first to frame hand-object interaction detection as the task of detecting hands, inferring contact states, and detecting manipulated objects, considered both egocentric and third-person vision images in which hands and objects are clearly visible.
This task formulation was later extended in~\cite{cheng2023towards} adding object and hands segmentation, secondary objects, and grasp type prediction.
Similar investigations have been also performed considering purely egocentric vision scenarios.
The authors of~\cite{lu2021egocentric} proposed an architecture leveraging specific egocentric cues such as hand poses and object masks.
The authors of~\cite{Ragusa2021TheMD} defined Egocentric Human-Object Interaction (EHOI) detection as the task of identifying manipulated objects and predicting interaction verbs for each of them. 
Hand-object interaction detection has also been considered in an industrial scenario in which manipulated object classes are known beforehand~\cite{leonardi2022egocentric,ragusa2024enigma}. Other investigations considered different tasks related to hand-object interaction understanding such as State Change Object Interaction Detection~\cite{Grauman2021Ego4DAT} or active object detection~\cite{Fu2021SequentialDF}. 
Previous investigations assumed different task formulations, which makes it hard to compare methods and assess progress. Recently, the authors of~\cite{VISOR2022} provided a task formulation termed Hand-Object Segmentation (HOS) together with the EPIC-KITCHENS VISOR dataset, thus setting a standard benchmark for hand-object interaction detection in egocentric vision.
HOS consists in estimating the contact relation between the hands and the objects and segmenting them given a single RGB frame.
A similar formulation was proposed in~\cite{EgoHos_jianbo_eccv22} together with the EgoHOS dataset.
In this paper, we adopt the HOS formulation and the baseline model of~\cite{VISOR2022} to perform experiments on three datasets designed for hand-object interaction detection: VISOR~\cite{VISOR2022}, EgoHOS~\cite{EgoHos_jianbo_eccv22}, and ENIGMA-51~\cite{ragusa2024enigma}.

\noindent\textbf{Simulators for Synthetic Visual Data Generation}
Previous research introduced simulators to generate synthetic data that mimics the behavior of real-world agents, such as cars and robots. Some examples include CARLA~\cite{Dosovitskiy17}, Gibson~\cite{xiazamirhe2018gibsonenv, li2022igibson}, Habitat~\cite{habitat19iccv, szot2021habitat}, Omniverse~\cite{Omniverse_nvidia}, and Isaac Sim~\cite{isaacSim_nvidia_2021}. Additionally, with advancements in graphics, game engines 
%such as Grand Theft Auto (GTA) V\footnote{\url{https://www.rockstargames.com/it/gta-v}} 
have been exploited for synthetic data generation for tasks such as pedestrian detection and tracking in urban scenarios~\cite{fabbri21iccv,di2021learning}, and safety monitoring in construction sites~\cite{quattrocchi2023Outline_SAFER}. Other studies proposed simulators specifically crafted to represent human agents moving in the scene~\cite{orlando2020egocentric} and interacting with objects~\cite{ai2thor}. Specifically, \cite{ai2thor} provides accurate modelling of the physics of the world and object manipulation actions, but it does not model hands, which are not visible in the scene. 
While these works advanced the understanding of synthetic data for computer vision tasks, no prior work studied the use of synthetic data for egocentric hand-object interaction detection, mainly due to the challenges associated with accurate modelling of environments, objects, and grasping. Based on recent advances in these fields~\cite{wang2023dexgraspnet,ramakrishnan2021habitat}, we propose a novel data generation pipeline and develop a simulator allowing to obtain realistic and diverse images of hand-object interactions.

\noindent\textbf{Synthetic Data for Hand-Object Interaction Understanding}
Few previous efforts used synthetic data to address tasks related to understanding egocentric hand-object interactions. The authors of~\cite{Hasson2019LearningJR} introduced the ObMan dataset designed for the joint reconstruction of hands and manipulated objects. 
The authors of~\cite{Jian_2023_ICCV} introduced a dataset of hand-object interactions designed for hand-object affordance understanding. 
Recently, \cite{ye2023affordance} exploited diffusion models for generating hand-object interactions synthetic datasets. 
These works did not tackle hand-object interaction detection and did not consider the generation of data with fine-grained labels required to address the task 
To enable the exploitation of synthetic data in this domain, we introduce a new benchmark comprising real images, paired with photorealistic labeled synthetic images annotated with labels useful for hand-object detection, such as 2D bounding boxes, semantic segmentation masks, depth maps, and hand-object relations.

\noindent\textbf{Domain Adaptation}
Domain Adaptation (DA) techniques have gained significant attention in recent years~\cite{csurka2017domain, saito2018maximum, bousmalis2017unsupervised, zhuang2020comprehensive}. These methods rely on different strategies to reduce the domain gap between a source and a target domain, such as adversarial training~\cite{ganin2015unsupervised, tzeng2017adversarial}, transfer learning~\cite{ganin2015unsupervised, tzeng2017adversarial} and pseudo-labeling~\cite{tarvainen2017mean, liu2021unbiased, li2022cross, cai2019exploring, deng2021unbiased}. While domain adaptation has been extensively studied in egocentric vision for different tasks, such as Action Recognition~\cite{Plizzari2023, munro_multimodal_domain_2019}, Person Re-identification~\cite{choudhary2021domain}, Video Retrieval~\cite{Munro2021DomainAI}, Ego-Exo Adaptation~\cite{Li2021EgoExoTV} and Object Detection~\cite{pasqualino2021unsupervised}, a study on DA for the HOI detection task is missing.
Unlike standard object detection, HOI detection involves identifying hands and active objects and understanding the specific relations between them. 
Hand-object interactions can be significantly influenced by the variability in shapes and sizes of objects, the diverse poses of human hands as well as by different contexts, such as kitchens and industrial laboratories. 
We provide HOI-Synth, the first benchmark explicitly designed to support the study of domain adaptation for hand-object interaction detection.

% -----------------------
% THE HOI-SYNTH BENCHMARK 
% -----------------------

\section{The HOI-Synth Benchmark} \label{sec:benchmark}
 \begin{figure}[t]
     \centering
     \includegraphics[width=\linewidth]{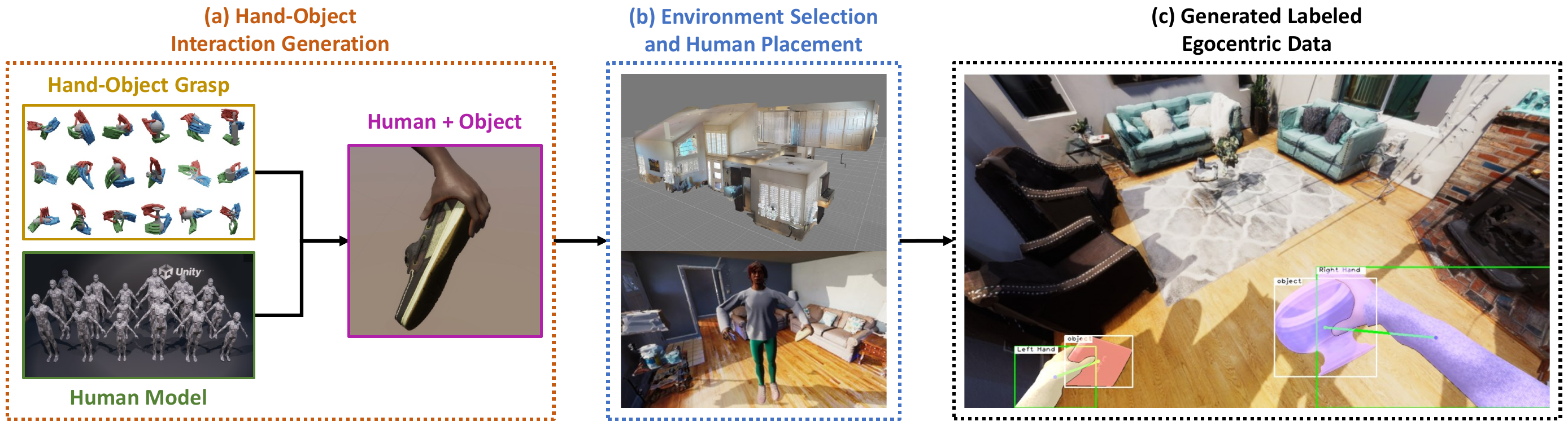}
     \caption{\textbf{The proposed data generation pipeline.} (a) An object-grasp pair is selected from DexGraspNet~\cite{wang2023dexgraspnet} 
and integrated with a randomly generated human model. (b) The human + object model is placed in an environment randomly selected from the Habitat-Matterport 3D dataset~\cite{ramakrishnan2021habitat}. (c) Egocentric data of hand-object interactions is generated and automatically labeled. Labels include bounding boxes and segmentation masks of hands and interacted objects, contact-state, and hand-object relations.}
     \label{fig:unity_tool}
 \end{figure}

In this section, we describe the HOI-Synth benchmark that we introduce to enable the study of synth-to-real domain adaptation for egocentric hand-object interaction detection. 
HOI-Synth is obtained by complementing three existing egocentric hand-object benchmarks with synthetic data, which is possible thanks to a novel data generation pipeline and a hand-object interaction simulator, which we release together with the benchmark to support future investigations on the use of synthetic data in egocentric vision.

\subsection{HOI-Synth Data Generation Pipeline and Simulator} \label{sec:simulator}
Figure~\ref{fig:unity_tool} shows a scheme of the proposed data generation pipeline, which is composed of three main steps. 
Our pipeline relies on state-of-the-art datasets and components to enable an accurate generation of egocentric images of hand-object interactions~\cite{wang2023dexgraspnet,unity_synthetichumans_2022,ramakrishnan2021habitat,shadowhand}.
We first select a random hand-object grasp from the DexGraspNet dataset~\cite{wang2023dexgraspnet}, which is fit to a randomly generated human model and integrated with the appropriate object mesh specified in the hand-object grasp~\cite{unity_synthetichumans_2022} (Figure~\ref{fig:unity_tool}-a).
We then select a random environment from the HM3D dataset~\cite{ramakrishnan2021habitat}  and place the human-object model in the environment (Figure~\ref{fig:unity_tool}-b).
We finally place a virtual camera at human eye level to capture the scene from the first-person point of view. 
For each generated interaction, the simulator annotates the bounding boxes and the segmentation masks of the hands and interacted objects, the hand contact state, as well as the hand-object relations (see Figure~\ref{fig:unity_tool}-c). 
We developed the pipeline in the Unity3D framework and implemented a hand-object interaction simulator, which will be publicly released to support future research on synthetic data generation for egocentric vision. The supplementary material reports details on the generation pipeline and visual examples of the generated data.

\subsection{Datasets}\label{sec:datasets}
The HOI-Synth benchmark extends three established datasets of egocentric images designed to study hand-object interaction detection, EPIC-KITCHENS VISOR~\cite{VISOR2022}, EgoHOS~\cite{EgoHos_jianbo_eccv22}, and ENIGMA-51~\cite{ragusa2024enigma}, with automatically labeled synthetic data obtained through the proposed generation pipeline. 

\noindent
{\textbf{EPIC-KITCHENS VISOR~\cite{VISOR2022}}} contains 36 hours of egocentric videos from EPIC-KITCHENS-100 \cite{Damen2021RESCALING}, including 32,857 training images and pixel-wise annotations for 42,787 hand-object relations. We complement this dataset with 30,259 synthetic images including 45,353 HOIs. 

\noindent
{\textbf{EgoHOS~\cite{EgoHos_jianbo_eccv22}}} 
includes 8,107 egocentric training images of HOIs sparsely sampled from videos belonging to EGO4D~\cite{Grauman2021Ego4DAT}, THU-READ~\cite{Tang2017ActionRI}, EPIC-KITCHENS~\cite{damen2018scaling}, and other egocentric videos of people playing escape rooms. 
The dataset is labeled with pixel-wise annotations of 13,659 hand-object relations. We complement this dataset with 8,107 synthetic images including 12,129 HOIs.

\noindent
{\textbf{ENIGMA-51~\cite{ragusa2024enigma}}} is an egocentric dataset of subjects following instructions to repair electrical boards in an industrial laboratory. The dataset contains 3,479 training images with pixel-wise annotations of 13,659 hand-object interactions. The dataset also provides 3D models of the manipulated objects and the industrial laboratory.
We complement this dataset with two sets of synthetic images: an in-domain set and an out-domain set. The in-domain set is generated using the 3D models of the environment and objects provided by the authors, thus obtaining synthetic images aligned to the real data. 
The out-domain set contains images of hand-object interactions in generic environments and with generic objects, akin to those generated to complement VISOR and EgoHOS.\footnote{\label{supp1}See the supplementary material for examples of in-domain and out-domain generated images and for additional details about architectures and training setups.} 

Table~\ref{tab:datasets} reports statistics of the training section of the HOI-Synth benchmark dataset, including the number of real and synthetic images, annotated hands, objects and HOIs. We use the official validation and test sets of the respective datasets for evaluation.

\begin{table}[t]
  \centering
    \caption{Statistics of the training sets considered in our HOI-Synth benchmark.}
  \resizebox{0.6\linewidth}{!}{
    \begin{tabular}{lrrrr}
      \hline
      \textbf{Dataset}                    & \textbf{Images} & \textbf{Hands} & \textbf{Objects} & \textbf{HOI} \\
      \hline               
      VISOR~\cite{VISOR2022}              & 32,857          & 52,906         & 42,785           & 42,787       \\
      Synthetic                           & 30,259          & 60,098         & 45,219           & 45,353       \\ 
      \hline 
      EgoHOS~\cite{EgoHos_jianbo_eccv22} & 8,107           & 15,015         & 11,393           & 13,659       \\ 
      Synthetic                           & 8,107           & 16,101         & 12,170           & 12,129       \\ 
      \hline
      ENIGMA-51~\cite{ragusa2024enigma} & 3,479           & 5,075          & 4,343            & 4,344        \\ 
      Synthetic-In-Domain                 & 16,773          & 25,444         & 16,637           & 16,773       \\     
      Synthetic-out-domain             & 20,321          & 40,135         & 27,499           & 27,370       \\ 
            
      \hline
    \end{tabular}
  }
  \label{tab:datasets}
\end{table}

% ---------------------------------
% EXPERIMENTAL ANALYSIS AND RESULTS
% ---------------------------------
\section{Experimental Analysis and Results} \label{sec:results}

We use \textit{VISOR HOS}~\cite{VISOR2022} as a baseline for our experiments. This method is based on the \textit{PointRend}~\cite{kirillov2020pointrend} instance segmentation network with the addition of three modules to detect the hand side, contact state (``contact'' or ``no contact''), and an offset vector that links the hand to the interacted object\footref{supp1}. We consider five different approaches to hand-object segmentation based on \textit{VISOR HOS}: 

\noindent
\textbf{Synthetic-Only} The \textit{VISOR HOS} model is trained using only synthetic data and tested directly on real data. Experiments with this approach aim to assess whether synthetic data can entirely replace real data. 

\noindent
\textbf{Unsupervised Domain Adaptation (UDA)} It replicates the \textit{VISOR HOS} architecture within the Adaptive Teacher (AT) unsupervised domain adaptation framework proposed in~\cite{li2022cross}. While AT was originally designed to tackle cross-domain object detection, we adapted it to perform HOI detection by adding modules to estimate hand side, contact state, offset vector, and segmentation masks. The model is hence trained using labeled synthetic data and unlabeled real data following an unsupervised domain adaptation scheme. This approach aims to assess whether current domain adaptation techniques in conjunction with high-quality labeled synthetic data allow avoiding labeling real data. 

\noindent
\textbf{Real-Only} It consists in training the \textit{VISOR HOS} model on labeled real data only. We experiment with different amounts of labeled real data to assess how much performance depends on the scale of training data when synthetic images are not available. This method provides baseline performance and corresponds to the standard fully supervised setup.

\noindent
\textbf{Synthetic + Real} It consists in pre-training the \textit{VISOR HOS} model on labeled synthetic data and fine-tuning it on labeled real data. Also in this case, we experiment with different amounts of labeled real data. These experiments aim to assess the potential of labeled synthetic data to reduce the amount of real data required for training, without any explicit synth-to-real adaptation.

\noindent
\textbf{Semi-Supervised Domain Adaptation (SSDA)} It consists in training the Adaptive Teacher model with labeled synthetic data, unlabeled real data, and a set of labeled real data. We experiment with different proportions of labeled real data (i.e., 10\%, 25\%, 50\%). To allow the Adaptive Teacher model to work in a semi-supervised regime, we merge the set of labeled synthetic data with the labeled real data. These experiments aim to assess whether synthetic data can improve results when only some real data are labeled.

\noindent
\textbf{Fully-Supervised Domain Adaptation (FSDA)}
We train the AT model merging labeled synthetic and all real data, while also performing domain adaptation. This approach aims to assess whether synthetic data can improve state-of-the-art results, even in the presence of large quantities of real labeled data.

\noindent
\textbf{Evaluation measures} Following~\cite{VISOR2022}, we evaluate performance using \textit{COCO Mask AP}~\cite{coco_dataset}. In particular, we adopted the 
Hand + Object (Overall) AP which assesses the correctness of the predicted hands and object bounding boxes of hands, the hand-state (contact vs. no contact) and the offset vector representing the relation between the hand and the active object.
We also break down performance using Mask APs measures evaluating specific aspects of the predictions: Hand (H), Hand + Side (H+S), Hand + Contact (H+C), and Object (O). 

% ---------------------------------
% VISOR TABLE
% ---------------------------------
\begin{table}[!t]
	\centering
        \caption{Results on the EPIC-KITCHENS VISOR validation set considering different real data settings available in training. Yellow rows indicate {\sethlcolor{yellow!20}\hl{baseline models}} in each configuration, while green rows highlight {\sethlcolor{customgreen!20}\hl{models trained with synthetic and real data}}. In each group, the \textbf{best results} are in bold, while the \underline{best results among the models trained with synthetic and real data} are underlined. \textcolor{blue}{Overall enhancements} are shown in blue, indicating improvements of the {\sethlcolor{customgreen!20}\hl{models}} trained with synthetic and real data over the {\sethlcolor{yellow!20}\hl{baseline}.}}
		
	\begin{minipage}{.9\linewidth} 
		\centering
		\textbf{a) Unsupervised Setting} \\
		\resizebox{\linewidth}{!}{
			    
			\begin{tabular}{c|c|c|cccc}
				\hline
				\textbf{\% Real Labeled Data} & \textbf{Approach}              & \textbf{Overall}                          & H                                         & H+S                                       & H+C                                       & O                                        \\
				\hline
				\multirow{2}{*}{0\%}   & \cellcolor{yellow!20} Synthetic-Only     & \cellcolor{yellow!20} 09.88 
				& \cellcolor{yellow!20} 28.41                           & \cellcolor{yellow!20} 24.89                           & \cellcolor{yellow!20} 08.64                           & \cellcolor{yellow!20} 01.23                                                     \\ 
				                              & \cellcolor{customgreen!15} UDA & \cellcolor{customgreen!15} \textbf{33.33} & \cellcolor{customgreen!15} \textbf{80.16} & \cellcolor{customgreen!15} \textbf{65.98} & \cellcolor{customgreen!15} \textbf{33.47} & \cellcolor{customgreen!15} \textbf{8.35} \\ 
				\hline
				\multicolumn{2}{c|}{Absolute Improvement}   & $\textcolor{blue}{\textbf{+23.45}}$& $+51.75$ & $+41.09$ & $+24.83$ & $+7.12$  \\

				\hline
							      
			\end{tabular} 
		}
	\end{minipage}

	\vspace{1em}
	        
	\begin{minipage}{.9\linewidth} 
		\centering
		\textbf{b) Semi-supervised Setting} \\
		\resizebox{\linewidth}{!}{
			       
			\begin{tabular}{c|c|c|cccc}
				\hline
				\textbf{\% Real Labeled Data} & \textbf{Approach}                         & \textbf{Overall}                                      & H                                                     & H+S                                          & H+C                                                   & O                                                     \\
				\hline
				\multirow{3}{*}{\makecell{10\% \\ (3,286 images)}}  & \cellcolor{yellow!20} Real-Only          
				& \cellcolor{yellow!20} 38.55 & \cellcolor{yellow!20} 87.45                           & \cellcolor{yellow!20} \textbf{83.27}                  & \cellcolor{yellow!20} 51.98                           & \cellcolor{yellow!20} 19.47                                                     \\ 
				                              & \cellcolor{customgreen!15} Synthetic+Real & \cellcolor{customgreen!15} 37.62                      & \cellcolor{customgreen!15} 86.39                      & \cellcolor{customgreen!15} \underline{82.85} & \cellcolor{customgreen!15} \textbf{\underline{52.25}} & \cellcolor{customgreen!15} \textbf{\underline{23.03}} \\ 
				                              & \cellcolor{customgreen!15} SSDA           & \cellcolor{customgreen!15} \textbf{\underline{44.22}} & \cellcolor{customgreen!15} \textbf{\underline{89.05}} & \cellcolor{customgreen!15}  80.77            & \cellcolor{customgreen!15} 46.83                      & \cellcolor{customgreen!15} 20.41                      \\ 
				\hline
				\multicolumn{2}{c|}{Absolute Improvement}  & $\textcolor{blue}{\textbf{+5.67}}$ & $+1.60$ & $-0.42$ & $+0.27$ & $+3.56$  \\
				
				% \multicolumn{2}{c|}{Relative Improvement}  & $\textcolor{blue}{\textbf{+14.71}}$ & $+1.82$ & $-0.50$ & $+0.52$ & $+18.28$  \\
							
				\hline
				\multirow{3}{*}{\makecell{25\% \\ (8,215 images)}}  & \cellcolor{yellow!20} Real-Only           
				& \cellcolor{yellow!20} 37.90
				& \cellcolor{yellow!20} 90.14                           & \cellcolor{yellow!20} \textbf{85.66}                  & \cellcolor{yellow!20} 53.99                           & \cellcolor{yellow!20} 17.85                                                     \\ 
				& \cellcolor{customgreen!15} Synthetic+Real 
				& \cellcolor{customgreen!15} 38.19 
				& \cellcolor{customgreen!15} 89.98                      & \cellcolor{customgreen!15} \underline{84.67}          & \cellcolor{customgreen!15} \textbf{\underline{55.88}} & \cellcolor{customgreen!15} 18.49                                          \\ 
				& \cellcolor{customgreen!15} SSDA   & \cellcolor{customgreen!15} \textbf{\underline{45.55}}        
				& \cellcolor{customgreen!15} \textbf{\underline{90.37}} & \cellcolor{customgreen!15} 84.42                      & \cellcolor{customgreen!15} 52.59                      & \cellcolor{customgreen!15} \textbf{\underline{22.15}}  \\ 
				\hline
				\multicolumn{2}{c|}{Absolute Improvement} & $\textcolor{blue}{\textbf{+7.65}}$  & $+0.23$ & $-0.99$ & $+1.89$ & $+4.30$ \\
				% \multicolumn{2}{c|}{Relative Improvement} & $\textcolor{blue}{\textbf{+20.18}}$  & $+0.25$ & $-1.16$ & $+7.88$ & $+24.09$ \\
				\hline
				\multirow{3}{*}{\makecell{50\% \\ (16,429 images)}}  & \cellcolor{yellow!20} Real-Only   
				& \cellcolor{yellow!20} 38.15
				& \cellcolor{yellow!20} 91.16                           & \cellcolor{yellow!20} \textbf{86.05}                  & \cellcolor{yellow!20} 52.28                           & \cellcolor{yellow!20} 17.92                                                     \\ 
				& \cellcolor{customgreen!15} Synthetic+Real 
				& \cellcolor{customgreen!15} 43.52 
				& \cellcolor{customgreen!15} \textbf{\underline{91.34}} & \cellcolor{customgreen!15} \underline{85.85}          & \cellcolor{customgreen!15} 54.09                      & \cellcolor{customgreen!15} 19.06                                           \\
				& \cellcolor{customgreen!15} SSDA          
				& \cellcolor{customgreen!15} \textbf{\underline{46.47}}
				& \cellcolor{customgreen!15} 90.94                      & \cellcolor{customgreen!15} 85.73                      & \cellcolor{customgreen!15} \textbf{\underline{58.02}} & \cellcolor{customgreen!15} \textbf{\underline{23.49}} \\ 
				\hline
				\multicolumn{2}{c|}{Absolute Improvement} & $\textcolor{blue}{\textbf{+8.32}}$  & $+0.18$ & $-0.20$ & $+5.74$ & $+5.57$  \\
				% \multicolumn{2}{c|}{Relative Improvement} & $\textcolor{blue}{\textbf{+21.81}}$  & $+0.20$ & $-0.23$ & $+10.98$ & $+31.08$  \\
				   
				\hline  
			\end{tabular}
		}
	\end{minipage}

	\vspace{1em}
	       
	\begin{minipage}{.9\linewidth} 
		\centering
		\textbf{c) Fully-supervised Setting} \\
		\resizebox{\linewidth}{!}{
			       
			\begin{tabular}{c|c|c|cccc}
				\hline
				\textbf{\% Real Labeled Data} & \textbf{Approach} & \textbf{Overall} & H & H+S & H+C & O \\
				\hline
				\multirow{3}{*}{\makecell{100\% \\ (32,857 images)}}
				& \cellcolor{yellow!20} Real-Only 
				& \cellcolor{yellow!20} 45.33& \cellcolor{yellow!20} \textbf{92.25}                  & \cellcolor{yellow!20} 88.54                           & \cellcolor{yellow!20} \textbf{59.24}                  & \cellcolor{yellow!20} 24.23                                                      \\ 
				& \cellcolor{customgreen!15} Synthetic+Real 
				& \cellcolor{customgreen!15} 44.52& \cellcolor{customgreen!15} 91.45                      & \cellcolor{customgreen!15} \textbf{\underline{88.94}} & \cellcolor{customgreen!15} 56.55                      & \cellcolor{customgreen!15} \textbf{\underline{27.77}}                       \\
				& \cellcolor{customgreen!15} FSDA   & \cellcolor{customgreen!15} \textbf{\underline{46.48}}       
				& \cellcolor{customgreen!15} \underline{91.83}          & \cellcolor{customgreen!15} 87.65                      & \cellcolor{customgreen!15} \underline{57.63}          & \cellcolor{customgreen!15} 24.03                       \\
				\hline
				\multicolumn{2}{c|}{Absolute Improvement} & $\textcolor{blue}{\textbf{+1.15}}$ & $-0.42$  & $+0.40$ & $-1.61$ & $+3.54$  \\
				
				% \multicolumn{2}{c|}{Relative Improvement} & $\textcolor{blue}{\textbf{+2.54}}$ & $-0.45$  & $+0.45$ & $-2.71$ & $+14.60$  \\
				\hline
				    			      
			\end{tabular}
		}
	\end{minipage}	  
	\label{tab:hos_visor}
 
\end{table}

\subsection{Results on VISOR} \label{sec:visor_results}
Table~\ref{tab:hos_visor} shows the results on the validation set of EPIC-KITCHENS VISOR~\cite{VISOR2022}.\footnote{Note that, in our implementation, the results of the HOS model differ from those reported in~\cite{VISOR2022} because, for fair comparisons, we adopted a batch size of $4$, the largest batch size achievable with domain adaptation models in our configuration.}
When no real labeled data are considered (Table~\ref{tab:hos_visor}-a), training the model only on synthetic data leads to poor performance, with an overall AP of $9.88\%$, which is not comparable to results achieved in fully supervised settings ($45.33\%$ when all real data are considered, as shown by \textit{Real-Only} in Table~\ref{tab:hos_visor}-c).
This highlights that, despite the photorealism of state-of-the-art data generation pipelines, there is still a consistent gap between synthetic and real data.
Adopting the Unsupervised Domain Adaptation (UDA) settings significantly improves model performance across all the evaluation criteria compared to the \textit{Synthetic-Only} approach. We observe an absolute improvement of $+23.45\%$ for the Overall Mask AP. Improvements are noticeable also across the different breakdown Mask APs: $+51.75\%$, $+41.09\%$, $+24.83\%$ for hand-dependent APs, and $+7.12\%$ for Object AP. This confirms the usefulness of synthetic data when Unsupervised Domain Adaptation approaches are used to mitigate the synthetic-real domain gap.
When different percentages of real labeled data (i.e., $10\%, 25\%$ and $50\%$) are considered in the semi-supervised setting (Table~\ref{tab:hos_visor}-b), models trained on synthetic and real data either via pre-training (\textit{Synthetic+Real}) or semi-supervised domain adaptation (\textit{SSDA}) achieve consistently higher performance with respect to baselines trained only with real data. In particular, the \textit{SSDA} approach achieves improvements of $+5.67\%$, $+7.65\%$, and $+8.32\%$ considering the Overall Mask AP metric when $10\%$, $25\%$, and $50\%$ of labeled real data are considered respectively. Significant improvements are observed when considering Object Mask AP (O) (+3.59\%, +4.30\% and +5.57\%), and H+C Mask AP ($+0.27\%, +1.89\%$ and $+5.74\%$), highlighting that synthetic data enhances the detection of active objects and improves the prediction of the hand contact state. 
Results are comparable with respect to the H and H+S measures, with improvements in the $[-0.99, +1.6]$ range, due to the fact that there is less room for improvement in these measures (all numbers in the $80\%-90\%$ range) and that real-only tends to overfit to these sub-tasks, while reaching suboptimal overall results.
When we consider a Fully-supervised configuration (Table~\ref{tab:hos_visor}-c), \textit{FSDA} improves the results over \textit{Real-Only} by a $+1.15\%$ in Overall Mask AP and by a $+3.54\%$ in Object Mask AP, with comparable performance on H, H+S, and H+C, with improvements in the range $[-1.61, +0.4]$.
It is worth noting that with as little as $25\%$ labeled real training data, which corresponds to only 8,215 images, \textit{SSDA} achieves an Overall Mask AP value of $45.55\%$, i.e., a $+7.65\%$ with respect to \textit{Real-Only} and even a $+0.22\%$ with respect to the fully supervised baseline trained with $100\%$ of all labeled real data (32,857 images).

\begin{table}[t]
    \begin{minipage}[b]{0.49\linewidth}
        \centering
        \caption{Results of different semi-supervised adaptation approaches trained with synthetic data and $25\%$ EPIC-KITCHENS VISOR labeled training data.}
        \resizebox{\linewidth}{!}{
            \begin{tabular}{l|c|cccc}
                \hline
                \textbf{Method}                     & \textbf{Overall}  & H                 & H+S               & H+C               & O                 \\
                \hline               
                Synthetic + Real                    & 38.19             & 89.98             & \textbf{84.67}    & 55.88             & 18.49             \\
                \hline
                MT~\cite{tarvainen2017mean}         & 43.69             & 88.78             & 84.40             & \textbf{60.94}    & 21.89             \\ 
                MT+GRL~\cite{ganin2015unsupervised} & 43.97             & 88.64             & 84.27             & \underline{58.21} & 21.82             \\ 
                UT~\cite{liu2021unbiased}           & \underline{44.32} & \textbf{90.60}    & \underline{84.49} & 52.55             & \underline{22.11} \\ 
                AT~\cite{li2022cross}               & \textbf{45.55}    & \underline{90.37} & 84.42             & 52.59             & \textbf{22.15}    \\ 
                \hline
            \end{tabular}
        }
        \label{tab:comparison_da_visor}
    \end{minipage}\hfill%
    \begin{minipage}[b]{0.47\linewidth}
        \centering
        \includegraphics[width=\linewidth]{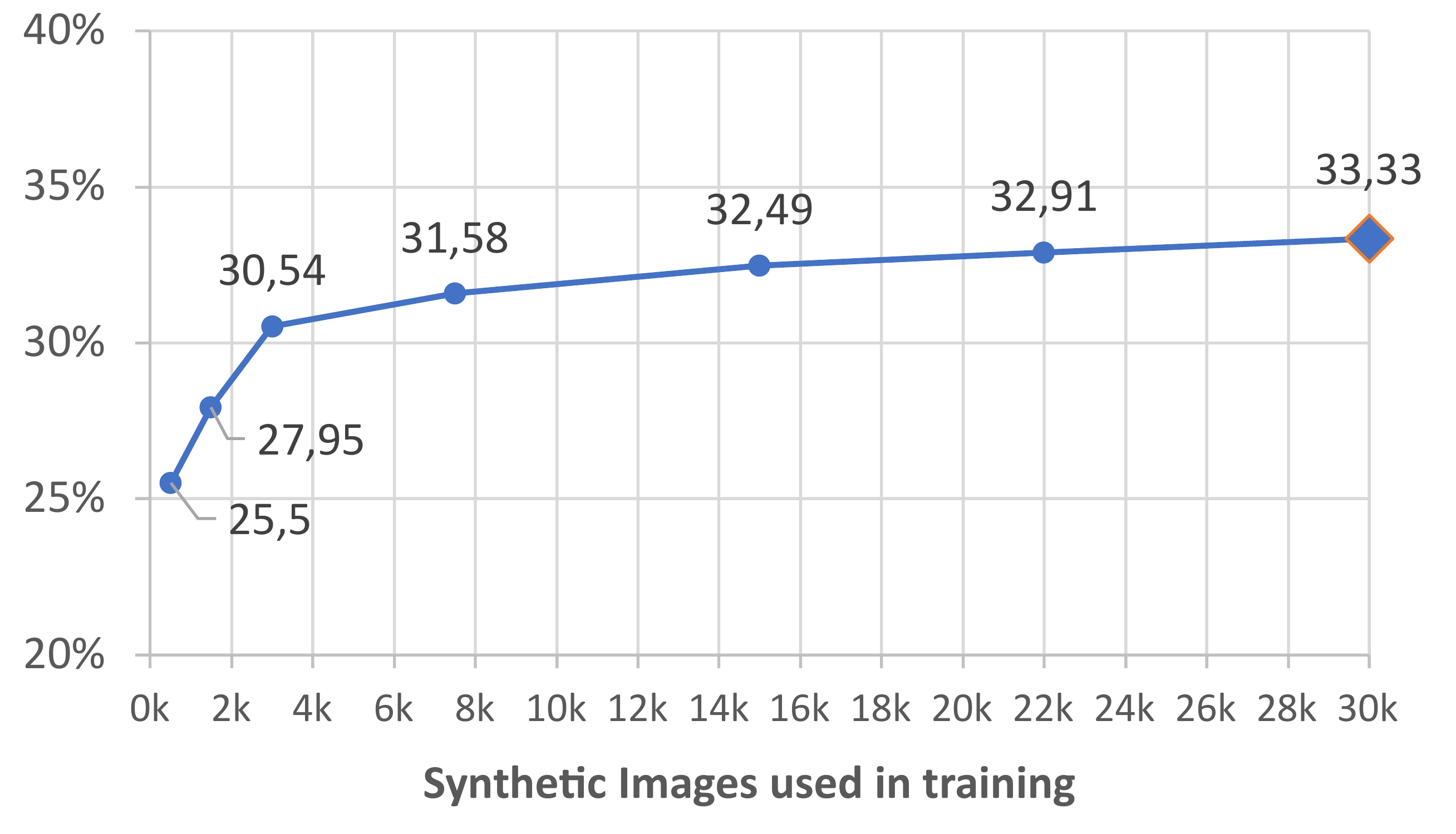}
        \captionof{figure}{UDA Overall AP on VISOR for different amounts of synthetic data.}
        \label{fig:synth}
    \end{minipage}%
\end{table}

\noindent\textbf{Benchmark of Domain Adaptation Approaches} In Table~\ref{tab:comparison_da_visor}, we compare the performance of different choices of semi-supervised domain adaptation approaches on EPIC-KITCHENS VISOR when $25\%$ of real labeled data are considered.
We choose this setup as a challenging benchmark for semi-supervised domain adaptation when the amount of labeled real data is limited. We considered the following methods: \textit{Mean Teacher~\cite{tarvainen2017mean}} (MT), \textit{Mean Teacher + Adversarial Loss~\cite{ganin2015unsupervised}} (MT+GRL), \textit{Unbiased Teacher~\cite{liu2021unbiased}} (UT), \textit{Adaptive Teacher~\cite{li2022cross}} (AT). In all cases, we adapted the methods to perform HOS by including appropriate layers for the prediction of hand side, contact state, offset vector, and segmentation masks. Methods are compared to the \textit{Synthetic + Real} baseline. 
\textit{MT} and \textit{MT+GRL} obtained the worst and second-worst results in the \textit{Overall AP} measure ($43.69\%$ and $43.97\%$), but achieved the best and second best results according to the \textit{AP Hand+Contact} measure ($60.94\%$ and $58.21\%$). \textit{UT} obtains the second best result in the \textit{Overall AP} ($44.32\%$), but also best result in the \textit{AP Hand} metric ($90.60\%$). It's worth noting that \textit{UT} outperforms the second-best results of \textit{AT} by a thin margin ($+0.23\%$) considering the \textit{Hand AP}, while \textit{AT} excels in other breakdown metrics.
Finally, \textit{AT} outperforms competitors in terms of the \textit{Overall AP}, surpassing UT by +$1.23\%$. Additionally, when considering the \textit{AP Object} measure, \textit{AT} achieves the best result of $22.15\%$. 
Results confirm the superior performance of the selected \textit{AT} method according to the \textit{Overall AP} performance measure with respect to the other domain adaptation strategies, despite not always achieving best results in breakdown metrics. This also suggests space for improvement in the proposed benchmark.\\
\noindent\textbf{Scale of Synthesized Data} Given that synthesized images can be easily generated at low cost and in large quantities using the proposed tool, we endeavoured to determine the scale of synthesized data required to maximize or plateau model performance.
To address this issue, we trained our UDA approach on VISOR using different amounts of labelled synthetic data and $25\%$ of real labeled data. Results reported in Figure~\ref{fig:synth} show how the model benefits from integrating additional quantities of synthetic data, approaching a plateau between 22k and 30k training images.

% ---------------------------------
% EGOHOS TABLE
% ---------------------------------
\begin{table}[!t]
  \centering
  \caption{Results on the EgoHOS~\cite{EgoHos_jianbo_eccv22} test set.} %See the caption of Table~\ref{tab:hos_visor} for a description of color coding. }
  
  \begin{minipage}{.9\linewidth} 
      \centering
     \textbf{a) Unsupervised Setting} \\
  
  \resizebox{\linewidth}{!}{
    \begin{tabular}{c|c|c|cccc}
      \hline
      \textbf{\% Real Labeled Data} & \textbf{Approach}     & \textbf{Overall}                    & H                                         & H+S                                    & H+C                                 & O                                                                        \\
      \hline
      \multirow{2}{*}{0\%}                     & \cellcolor{yellow!20} 
      Synthetic-Only    & \cellcolor{yellow!20} 07.16  & \cellcolor{yellow!20} 18.25                           & \cellcolor{yellow!20} 15.93                           & \cellcolor{yellow!20} 05.33                           & \cellcolor{yellow!20} 01.24                                                      \\ 
                                    & \cellcolor{customgreen!15} UDA    
                                     & \cellcolor{customgreen!15} \textbf{28.16}
                                    & \cellcolor{customgreen!15} \textbf{70.30}             & \cellcolor{customgreen!15} \textbf{59.21}             & \cellcolor{customgreen!15} \textbf{20.84}             & \cellcolor{customgreen!15} \textbf{09.65}                         \\ 
      \hline
      \multicolumn{2}{c|}{Absolute Improvement}  & $\textcolor{blue}{\textbf{+21.00}}$ & $+52.05$ & $+43.28$ & $+15.51$ & $+8.41$  \\ 
      % \multicolumn{2}{c|}{Relative Improvement}  & $\textcolor{blue}{\textbf{+293.30}}$ & $+285.20$ & $+271.69$ & $+290.99$ & $+678.23$  \\
      \hline
\end{tabular}
      }
      \end{minipage}
      \vspace{1em}
      
      \begin{minipage}{.9\linewidth} 
      \centering
     \textbf{b) Semi-supervised Setting} \\
  
  \resizebox{\linewidth}{!}{
  \begin{tabular}{c|c|c|cccc}

  \hline
      \textbf{\% Real Labeled Data} & \textbf{Approach}     & \textbf{Overall}                    & H                                         & H+S                                    & H+C                                 & O                                                                         \\
      \hline

\multirow{3}{*}{\makecell{10\% \\ (857 images)}}          	& \cellcolor{yellow!20} Real-Only   
 & \cellcolor{yellow!20} 28.44   
& \cellcolor{yellow!20} 76.28                           & \cellcolor{yellow!20} 68.92                           & \cellcolor{yellow!20} 35.84                           & \cellcolor{yellow!20} 16.59                                                   \\ 
																		& \cellcolor{customgreen!15} Synthetic+Real 
                  & \cellcolor{customgreen!15} 28.74
                  & \cellcolor{customgreen!15} 77.15                      & \cellcolor{customgreen!15} 71.64                      & \cellcolor{customgreen!15} 39.25                      & \cellcolor{customgreen!15} 17.33                                             \\ 
																		& \cellcolor{customgreen!15} SSDA 
                   & \cellcolor{customgreen!15} \textbf{\underline{36.68}}
                  & \cellcolor{customgreen!15} \textbf{\underline{83.25}} & \cellcolor{customgreen!15} \textbf{\underline{73.72}} & \cellcolor{customgreen!15} \textbf{\underline{47.20}} & \cellcolor{customgreen!15}  \textbf{\underline{22.40}} \\ 
			\hline
			\multicolumn{2}{c|}{Absolute Improvement}  
   & $\textcolor{blue}{\textbf{+8.24}}$
   & $+6.97$ &  $+4.80$   & $+11.36$  & $+5.81$  \\
			
			% \multicolumn{2}{c|}{Relative Improvement}  
   % & $\textcolor{blue}{\textbf{+28.97}}$
   % & $+9.14$ &  $+6.96$   & $+31.70$  & $+35.02$  \\
			\hline

      \multirow{3}{*}{\makecell{25\% \\ (2,026 images)}}         & \cellcolor{yellow!20} Real-Only       & \cellcolor{yellow!20} 33.73     & \cellcolor{yellow!20} 78.94                           & \cellcolor{yellow!20} 70.62                           & \cellcolor{yellow!20} 41.67                           & \cellcolor{yellow!20} 21.83                                                     \\ 
                                    & \cellcolor{customgreen!15} Synthetic+Real
                                    & \cellcolor{customgreen!15} 33.78
                                    & \cellcolor{customgreen!15} 79.60                      & \cellcolor{customgreen!15} 71.61                      & \cellcolor{customgreen!15} 46.11                      & \cellcolor{customgreen!15} 19.87                                            \\ 
                                    & \cellcolor{customgreen!15} SSDA     
                                    & \cellcolor{customgreen!15} \textbf{\underline{37.16}}
                                    & \cellcolor{customgreen!15} \textbf{\underline{83.79}} & \cellcolor{customgreen!15} \textbf{\underline{74.28}} & \cellcolor{customgreen!15} \textbf{\underline{49.00}} & \cellcolor{customgreen!15}  \textbf{\underline{23.82}}  \\ 
      \hline
      \multicolumn{2}{l|}{Absolute Improvement}  
       & $\textcolor{blue}{\textbf{+3.43}}$
      & $+4.85$ &  $+3.66$   & $+7.33$  & $+1.99$ \\
      % \multicolumn{2}{l|}{Improvement as relative change}  
      %  & $\textcolor{blue}{\textbf{+10.17}}$
      % & $+6.14$ &  $+5.18$   & $+17.59$  & $+9.12$ \\
      \hline
        \multirow{3}{*}{\makecell{50\% \\ (4,379 images)}}  		& \cellcolor{yellow!20} Real-Only		
         & \cellcolor{yellow!20} 36.30  & \cellcolor{yellow!20} 81.82                           & \cellcolor{yellow!20} 73.63                           & \cellcolor{yellow!20} 47.27                           & \cellcolor{yellow!20} 25.73                                                   \\ 
																		& \cellcolor{customgreen!15} Synthetic+Real
                  & \cellcolor{customgreen!15} 34.30 
                  & \cellcolor{customgreen!15} 82.54                      & \cellcolor{customgreen!15} 74.03                      & \cellcolor{customgreen!15} 47.92                      & \cellcolor{customgreen!15} 23.47                                           \\
																		& \cellcolor{customgreen!15} SSDA 
                  & \cellcolor{customgreen!15} \textbf{\underline{39.85}}	
                  & \cellcolor{customgreen!15} \textbf{\underline{85.17}}						& \cellcolor{customgreen!15} \textbf{\underline{76.80}}						& \cellcolor{customgreen!15} \textbf{\underline{52.58}}						& \cellcolor{customgreen!15} \textbf{\underline{26.90}}											\\ 
			\hline
			\multicolumn{2}{c|}{Absolute Improvement}  
    & $\textcolor{blue}{\textbf{+3.55}}$
   & $+3.97$ &  $+3.17$   & $+5.31$  & $+1.17$ \\
% \multicolumn{2}{c|}{Relative Improvement}  
%     & $\textcolor{blue}{\textbf{+9.78}}$
%    & $+4.85$ &  $+4.30$   & $+11.23$  & $+4.55$ \\
   
			\hline

      \end{tabular}
      }
      \end{minipage}
       \vspace{1em}
       
      \begin{minipage}{.9\linewidth} 
      \centering
     \textbf{c) Fully-supervised Setting} \\
  
  \resizebox{\linewidth}{!}{
      \begin{tabular}{c|c|c|cccc}
      \hline
      \textbf{\% Real Labeled Data} & \textbf{Approach}     & \textbf{Overall}                    & H                                         & H+S                                    & H+C                                 & O                                                                         \\
      \hline
      \multirow{3}{*}{\makecell{100\% \\ (8,758 images)}}        & \cellcolor{yellow!20} Real-Only    & \cellcolor{yellow!20} 36.16        & \cellcolor{yellow!20} 84.39                           & \cellcolor{yellow!20} 76.24                           & \cellcolor{yellow!20} 51.81                           & \cellcolor{yellow!20} 26.46                                                     \\ 
                                    & \cellcolor{customgreen!15} Synthetic+Real
                                    & \cellcolor{customgreen!15} 34.68
                                    & \cellcolor{customgreen!15} 84.56                      & \cellcolor{customgreen!15} 71.56                      & \cellcolor{customgreen!15} 49.72                      & \cellcolor{customgreen!15} 23.16                                            \\
                                    & \cellcolor{customgreen!15} FSDA    
                                    & \cellcolor{customgreen!15} \textbf{\underline{39.61}}
                                    & \cellcolor{customgreen!15} \textbf{\underline{85.58}} & \cellcolor{customgreen!15} \textbf{\underline{76.80}} & \cellcolor{customgreen!15} \textbf{\underline{51.99}} & \cellcolor{customgreen!15} \textbf{\underline{27.05}}  \\
      \hline
      \multicolumn{2}{c|}{Absolute Improvement}
      & $\textcolor{blue}{\textbf{+3.45}}$
      & $+1.19$  & $+0.56$  & $+0.18$  & $+0.59$   \\
      % \multicolumn{2}{c|}{Relative Improvement}
      % & $\textcolor{blue}{\textbf{+9.54}}$
      % & $+1.41$  & $+0.73$  & $+0.35$  & $+2.23$   \\
      \hline
    \end{tabular}
  }
  \end{minipage}
  \label{tab:hos_egohos}
\end{table}

\subsection{Results on EgoHOS} \label{sec:egohos_results}
Table~\ref{tab:hos_egohos} reports the results on the test set of EgoHOS~\cite{EgoHos_jianbo_eccv22}. 
Also in this case, using only synthetic data (Table~\ref{tab:hos_egohos}-a) does not allow to achieve satisfactory performance, highlighting the existence of a domain gap between real and synthetic data. Indeed, synthetic-only achieves an Overall AP of $7.16\%$, a $\sim 20\%$ drop with respect to a fully supervised baseline trained on $100\%$ labeled real data  (Table~\ref{tab:hos_egohos}-c).  \textit{UDA} significantly improves over \textit{Synthetic-Only}, achieving +$21.00\%$ on the Overall Mask AP, and improvements of  $+52.05\%$, $+43.28\%$, $+15.51\%$, and $+8.41\%$ across the breakdown metrics (H, H+S, H+C and O).
In the semi-supervised settings (Table~\ref{tab:hos_egohos}-b), both \textit{Synthetic+Real} and \textit{SSDA} improve over \textit{Real-Only}, with major improvements obtained by SSDA, which obtains the best \textit{Overall APs}, surpassing \textit{Real-Only} by $+8.24\%$, $+3.43\%$ and $+3.55\%$ when 10\%, 25\% and 50\% amounts of real labeled data are considered.  In the fully supervised settings (Table~\ref{tab:hos_egohos}-c), \textit{FSDA} consistently achieves best results across all measures, surpassing \textit{Real-Only} by $+3.45\%$ in \textit{Overall AP}. Notably, \textit{SSDA} trained with only $25\%$ real labeled data,  corresponding to 2,026 images, obtains an improvement of $+0.52\%$ with respect to \textit{Real-Only} trained with $100\%$ real labeled data (8,758 images) according to the \textit{Overall AP} measure.
These results confirm the effectiveness of synthetic data in reducing the need for real labeled data, and as a way to improve performance over standard fully supervised methods.

% ---------------------------------
% ENIGMA TABLE
% ---------------------------------
\begin{table}[t]
    \centering
    \caption{Results on the ENIGMA-51~\cite{ragusa2024enigma} test set.} 
    \begin{minipage}{.9\linewidth} 
        \centering
        \textbf{a) Unsupervised Setting}
        \resizebox{\linewidth}{!}{
                                                    
            \begin{tabular}{c|c|c|c|cccc}
                \hline
                \textbf{\% Real Labeled Data} & \textbf{Approach}                    & \textbf{In-domain}                    & \textbf{Overall}                          & H            & H+S                                       & H+C                                       & O                                         \\
                \hline
                \multirow{4}{*}{0\%}   
                
                  & \cellcolor{yellow!20} Synthetic-Only & \cellcolor{yellow!20}     & \cellcolor{yellow!20} 00.21 & \cellcolor{yellow!20} 01.07               & \cellcolor{yellow!20} 00.11               & \cellcolor{yellow!20} 00.03               & \cellcolor{yellow!20} 00.99    \\

                  & \cellcolor{yellow!20} Synthetic-Only & \cellcolor{yellow!20} \checkmark      & \cellcolor{yellow!20} 12.85               & \cellcolor{yellow!20} 56.05               & \cellcolor{yellow!20} 35.14               & \cellcolor{yellow!20} 15.24               & \cellcolor{yellow!20} 4.79                \\

                  & \cellcolor{customgreen!15} UDA       &  \cellcolor{customgreen!15}                                     & \cellcolor{customgreen!15} 6.87           & \cellcolor{customgreen!15} 42.81          & \cellcolor{customgreen!15} 14.52          & \cellcolor{customgreen!15} 7.97           & \cellcolor{customgreen!15} 3.29  \\

                  & \cellcolor{customgreen!15} UDA       & \cellcolor{customgreen!15} \checkmark & \cellcolor{customgreen!15} \textbf{\underline{34.78}} & \cellcolor{customgreen!15} \textbf{\underline{78.83}} & \cellcolor{customgreen!15} \textbf{\underline{70.91}} & \cellcolor{customgreen!15} \textbf{\underline{28.14}} & \cellcolor{customgreen!15} \textbf{\underline{25.84}} \\ \hline
                  
                \multicolumn{3}{c|}{Absolute Improvement}   & $\textcolor{blue}{\textbf{+21.93}}$       & $+22.78$  & $+35.77$  & $+12.90$  & $+21.05$  \\
                %\multicolumn{3}{c|}{Relative Improvement}   & $\textcolor{blue}{\textbf{+170.66}}$      & $+40.64$  & $+101.79$ & $+84.64$  & $+439.46$ \\
                \hline
                                                                                              
            \end{tabular} 
        }
    \end{minipage}
        
    \vspace{1em}
        
    \begin{minipage}{.9\linewidth} 
         \centering
        \textbf{b) Semi-supervised Setting} \\
        \resizebox{\linewidth}{!}{
       
        \begin{tabular}{c|c|c|c|cccc}
            
            %------------------10%
            \hline
            \textbf{\% Real Labeled Data} & \textbf{Approach}        & \textbf{In-domain}                 & \textbf{Overall}                                      & H                                                     & H+S                                                    & H+C                                                   & O                                                     \\
            
            \hline
            \multirow{3}{*}{\makecell{10\% \\ (347 images)}}  
            
            & \cellcolor{yellow!20} Real-Only   & \cellcolor{yellow!20} \checkmark         & \cellcolor{yellow!20} 45.39 & \cellcolor{yellow!20} 81.25                           & \cellcolor{yellow!20} 76.22                  & \cellcolor{yellow!20} 37.96                           & \cellcolor{yellow!20} 39.53                                                    \\

            & \cellcolor{customgreen!15} SSDA   & \cellcolor{customgreen!15}    & \cellcolor{customgreen!15} \textbf{\underline{57.08}}  & \cellcolor{customgreen!15} \textbf{\underline{85.40}} & \cellcolor{customgreen!15} \textbf{\underline{78.62}} & \cellcolor{customgreen!15} \textbf{\underline{43.56}} & \cellcolor{customgreen!15} \textbf{\underline{46.97}} \\

            & \cellcolor{customgreen!15} SSDA   & \cellcolor{customgreen!15} \checkmark    & \cellcolor{customgreen!15} 56.69 & \cellcolor{customgreen!15} 84.58 & \cellcolor{customgreen!15} 78.42 & \cellcolor{customgreen!15} 41.17                      & \cellcolor{customgreen!15} 46.50 \\ 
            \hline

            \multicolumn{3}{c|}{Absolute Improvement}  & $\textcolor{blue}{\textbf{+11.69}}$ & $+4.15$ & $+2.40$ & $+5.60$ & $+7.44$  \\
            %\multicolumn{3}{c|}{Relative Improvement}  & $\textcolor{blue}{\textbf{+25.75}}$ & $+5.11$ & $+3.15$ & $+14.75$ & $+18.82$  \\                        
            
            \hline

            %-----25%
         
            \multirow{3}{*}{\makecell{25\% \\ (870 images)}}  
            
            & \cellcolor{yellow!20} Real-Only & \cellcolor{yellow!20} \checkmark    & \cellcolor{yellow!20} 51.83 & \cellcolor{yellow!20} 82.95 & \cellcolor{yellow!20} 78.70                  & \cellcolor{yellow!20} 43.52                         & \cellcolor{yellow!20} 45.25   \\ 
            
            & \cellcolor{customgreen!15} SSDA & \cellcolor{customgreen!15}	& \cellcolor{customgreen!15} 58.17 & \cellcolor{customgreen!15} \textbf{\underline{84.99}}	& \cellcolor{customgreen!15} \textbf{\underline{80.41}}						& \cellcolor{customgreen!15} \textbf{\underline{46.31}}						& \cellcolor{customgreen!15} 49.34	\\

            & \cellcolor{customgreen!15} SSDA   & \cellcolor{customgreen!15} \checkmark & \cellcolor{customgreen!15} \textbf{\underline{59.48}}        
            & \cellcolor{customgreen!15} 84.85 & \cellcolor{customgreen!15} 80.30                      & \cellcolor{customgreen!15} 44.24                      & \cellcolor{customgreen!15} \textbf{\underline{\textbf{49.37}}}  \\

            \hline
            \multicolumn{3}{c|}{Absolute Improvement} & $\textcolor{blue}{\textbf{+7.65}}$  & $+2.04$ & $+1.71$ & $+2.79$ & $+4.12$ \\
            %\multicolumn{3}{c|}{Relative Improvement} & $\textcolor{blue}{\textbf{+14.76}}$  & $+2.45$ & $+2.17$ & $+6.41$ & $+9.11$ \\
            \hline

               %-----50%

                \multirow{3}{*}{\makecell{50\% \\ (1,739 images)}}  
                & \cellcolor{yellow!20} Real-Only & \cellcolor{yellow!20} \checkmark   & \cellcolor{yellow!20} 57.62
                & \cellcolor{yellow!20} 84.65                           & \cellcolor{yellow!20} 80.43                  & \cellcolor{yellow!20} 47.41                           & \cellcolor{yellow!20} 48.79                                                     \\

    			& \cellcolor{customgreen!15} SSDA           & \cellcolor{customgreen!15}     & \cellcolor{customgreen!15} \textbf{\underline{63.25}}			& \cellcolor{customgreen!15} \textbf{\underline{85.67}}	& \cellcolor{customgreen!15} 82.00						& \cellcolor{customgreen!15} \textbf{\underline{52.20}}						& \cellcolor{customgreen!15} \textbf{\underline{52.56}}		\\

    			& \cellcolor{customgreen!15} SSDA           & \cellcolor{customgreen!15} \checkmark     & \cellcolor{customgreen!15} 61.93			& \cellcolor{customgreen!15} 85.12	& \cellcolor{customgreen!15} \textbf{\underline{82.01}}						& \cellcolor{customgreen!15} 48.96						& \cellcolor{customgreen!15} 51.94		\\
                \hline
                
                \multicolumn{3}{c|}{Absolute Improvement} & $\textcolor{blue}{\textbf{+5.63}}$  & $+1.02$ & $+1.58$ & $+4.79$ & $+3.77$  \\
                %\multicolumn{3}{c|}{Relative Improvement} & $\textcolor{blue}{\textbf{+9.77}}$  & $+1.20$ & $+1.96$ & $+10.10$ & $+7.73$  \\
                   
                \hline  
    		\end{tabular}
      }
      \end{minipage}

       \vspace{1em}
       
        \begin{minipage}{.9\linewidth} 
         \centering
     \textbf{c) Fully-supervised Setting} \\
        \resizebox{\linewidth}{!}{
       
    \begin{tabular}{c|c|c|c|cccc}
    
            \hline
            \textbf{\% Real Labeled Data}  & \textbf{Approach} & \textbf{In-domain} & \textbf{Overall}                       & H                                         & H+S                                    & H+C                        & O                                                                         \\
            \hline
            \multirow{3}{*}{\makecell{100\% \\ (3,479 images)}}
            & \cellcolor{yellow!20} Real-Only & \cellcolor{yellow!20} \checkmark & \cellcolor{yellow!20} 63.84& \cellcolor{yellow!20} 85.01                  & \cellcolor{yellow!20} 81.05                           & \cellcolor{yellow!20} 52.32                  & \cellcolor{yellow!20} 51.35     \\

            & \cellcolor{customgreen!15} FSDA           & \cellcolor{customgreen!15}            & \cellcolor{customgreen!15} \textbf{\underline{64.41}} & \cellcolor{customgreen!15} \textbf{\underline{85.94}}                      & \cellcolor{customgreen!15} \textbf{\underline{82.91}} & \cellcolor{customgreen!15} \textbf{\underline{54.13}}                      & \cellcolor{customgreen!15} 52.50        \\

            & \cellcolor{customgreen!15} FSDA   & \cellcolor{customgreen!15} \checkmark & \cellcolor{customgreen!15} 64.20       
            & \cellcolor{customgreen!15} 85.37          & \cellcolor{customgreen!15} 82.45                      & \cellcolor{customgreen!15} 51.60          & \cellcolor{customgreen!15} \underline{\textbf{53.30}}                       \\
            
            \hline
            \multicolumn{3}{c|}{Absolute Improvement} & $\textcolor{blue}{\textbf{+0.57}}$ & $+0.93$  & $+1.86$ & $+1.81$ & $+1.95$  \\
            
            %\multicolumn{3}{c|}{Relative Improvement} & $\textcolor{blue}{\textbf{+0.89}}$ & $+1.09$  & $+2.29$ & $+3.46$ & $+3.80$  \\
            \hline
    			      
    		\end{tabular}
      }
      \end{minipage}

	%}
	\label{tab:hos_enigma}
\end{table}

\subsection{Results on ENIGMA-51} \label{sec:enigma_results}
Table~\ref{tab:hos_enigma} reports the results on the test set of ENIGMA-51~\cite{ragusa2024enigma}. In this case, we also compare performance when in-domain and out-domain synthetic data are used.
In the unsupervised settings (Table~\ref{tab:hos_enigma}-a), using only generic synthetic data leads to poor performances, confirming the domain gap between synthetic and real images also in this case.
Using in-domain synthetic real data greatly reduces the gap, with the Overall AP passing from $0.21\%$ to $12.85\%$.
The UDA approach improves results both when paired with generic and in-domain data. In the last case, UDA achieves an improvement of $+21.93\%$ in Overall AP, with improvements also in the breakdown APs: $+22.78\%$ (H), $+35.77\%$ (H+S), $+12.90\%$ (H+C) and $+21.05\%$ (O). 
The choice of the synthetic data source (in-domain vs. out-domain) is crucial, impacting the performance of models in this unsupervised setting. The UDA approach, trained with in-domain synthetic data, outperforms the same model trained with out-domain synthetic data by $27.91\%$ (\textit{Overall AP}). 
This result highlights that in-domain information contained in the generated images helps the detection of hand-object interactions when a specific domain is considered.
In the semi-supervised setting (Table~\ref{tab:hos_enigma}-b), \textit{SSDA} systematically outperforms the baseline obtaining gains across \textit{Overall Mask AP} of \textit{+11.69\%, +7.65\%} and \textit{+5.63\%} when \textit{10\%, 25\%} and \textit{50\%} real labeled data are considered.  
Gains are also observed across all measures in the fully supervised setting (Table~\ref{tab:hos_enigma}-c), e.g., with a $+0.57\%$ in overall AP, and a $+1.95\%$ in Object AP.
Interestingly, the choice of in-domain vs. out-domain synthetic data in the semi- and fully-supervised settings is not as crucial as in the case of unsupervised domain adaptation, with both data sources achieving overall similar performance across the different AP measures, suggesting that even small quantities of real labeled can bridge the gap between out-domain synthetic and real data, hence making the generation of in-domain data less critical.

\subsection{Qualitative results} \label{sec:qualitative_results}
\begin{figure}[t]
    \centering
    \includegraphics[width=0.8\linewidth]{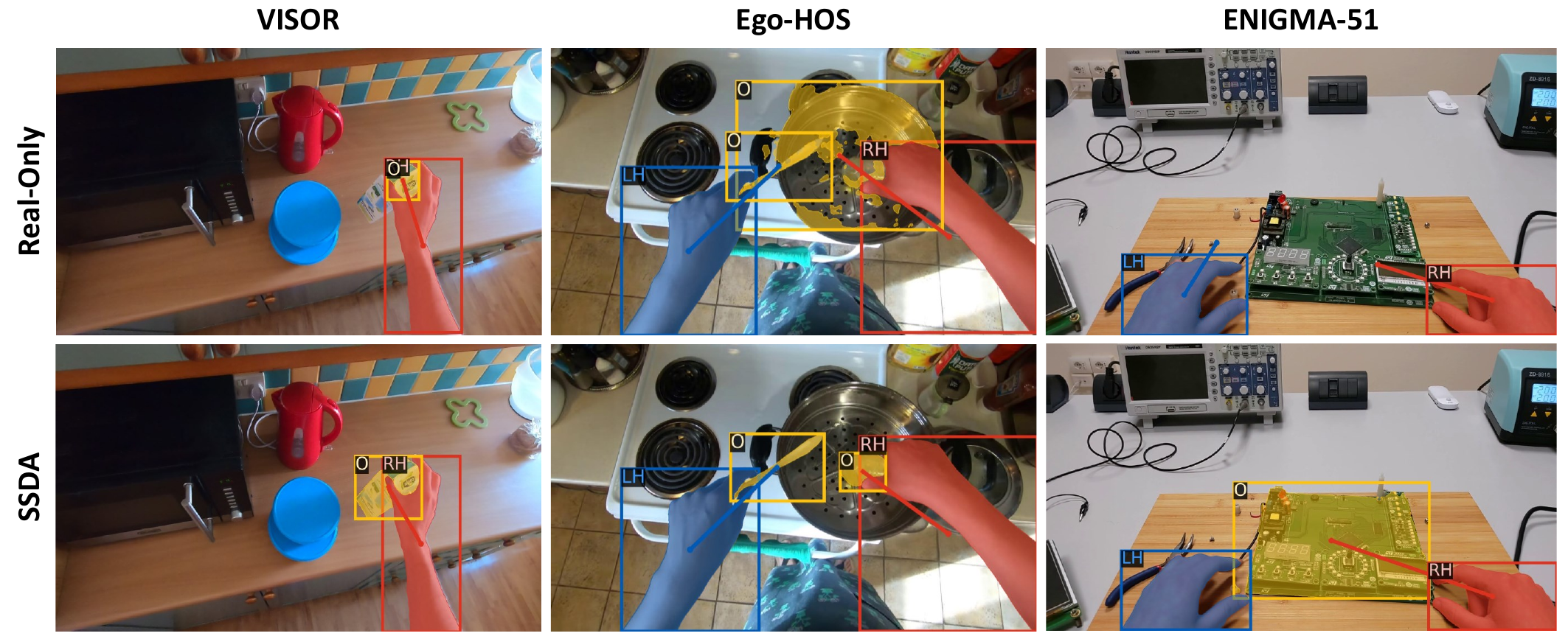}
    \caption{Qualitative examples of \textit{Real-Only} and \textit{SSDA} on the three datasets. \textit{SSDA} achieves better object segmentation and detection performance across the datasets.}
    \label{fig:qualitative_examples}
\end{figure}

Figure~\ref{fig:qualitative_examples} reports qualitative examples comparing \textit{SSDA} with respect to \textit{Real-Only} when $25\%$ real labeled data are considered.
In the VISOR example (first column), \textit{SSDA} (second row) obtains better object segmentation than \textit{Real-Only} (first row). This behavior can also be observed in the EgoHOS example (second column), where \textit{SSDA} better detects and segments the objects involved in the interaction. In the ENIGMA-51 example (third column), \textit{SSDA} detects and segments the interacted object, that was not detected by the \textit{Real-Only} approach.\footnote{Additional qualitative examples are reported in the supplementary material.}%\footnote{Additional qualitative examples are reported in supp. material.}. 

% ----------
% CONCLUSION
% ----------

\section{Discussion and Conclusion}\label{sec:conclusion}
With the proposed analysis we aimed to address several questions.\\
\textit{Is there a gap between synthetic and real data? Where does it originate? How can it be reduced?}
Despite progress in realistic data generation, a gap remains between synthetic and real data. Our analysis offers insights into the extent of such gap, which is in the order of $30\%-40\%$ depending on the dataset. In the context of VISOR, the estimated gap ($35.45\%$) is narrowed by unsupervised domain adaptation to $12.00\%$ and further shrunk to $1.11\%$ adopting semi-supervised domain adaptation strategies. Similar considerations can be made for the other datasets. We suggest this gap is caused by the photo-realism of generated synthetic data, the diversity of context-aware characteristics (as shown by results with in/out-domain synthetic data) and hand-object interactions. \\
\textit{Can synthetic data entirely replace real data?}
Our study suggests that synthetic data cannot yet entirely replace real data for egocentric hand-object interaction detection, with synthetic-only baselines achieving poor results in all scenarios.\\
\textit{Can synthetic data enable training in the presence of unlabeled real data?}
While synthetic data cannot entirely replace real data, we show that it greatly improves models' performance in the presence of unlabeled real data. Indeed, significant gains are obtained by UDA across all scenarios, when compared to a synthetic-only baseline, while the gap with respect to fully supervised baselines is narrowed. For instance, in the VISOR dataset, UDA obtained a $+23.45\%$ improvement with respect to real-only in Overall AP, obtaining a score of $33.33\%$, about $10\%$ smaller than the fully supervised baseline trained on real data.\\
\textit{Can synthetic data increase efficiency when few real data are labeled?}
When different amounts of real labeled data are exploited together with synthetic data, SSDA and FSDA models obtain improvements in \textit{Overall AP} over baselines trained on real data only in the considered benchmark. Notably, the performance gap diminishes as the quantity of real data increases: from +23.45\% (0\% of real data) to +1.15\% (100\% of real data) in VISOR, from +21.00\% (0\% of real data) to +3.45\% (100\% of real data) in EgoHOS and from +21.93\% (0\% of real data) to +2.33\% (100\% of real data) for ENIGMA-51. These results highlight the effectiveness of using synthetic data when real labeled data are scarce.\\
\textit{What scale of synthetic data is needed}
Our findings reveal that models benefit from large quantities of synthetic data. For instance, in the context of VISOR, a plateau is reached when 22K-30K synthetic images are included for training.\\
\textit{Is in-domain synthetic data beneficial?}
Our analysis shows that in-domain data is highly beneficial in unsupervised settings, where it helps narrow down the domain gap. For instance, in the ENIGMA-51 dataset, using in-domain synthetic data only allows to obtain an overall AP of $12.85$, about $+10\%$ with respect to out-domain data. With UDA, performance jumps to $34.78\%$, a major increase.
With few real labeled data, choice of in-domain data is less crucial, with models achieving comparable performance, regardless of the training data source.

We hope that our analysis will inform future application and model developments and that the release of the HOI-Synth benchmark and data generation pipeline will support future research in this field.

\section*{Acknowledgments}
This research has been supported by the project Future Artificial Intelligence Research (FAIR) – PNRR MUR Cod. PE0000013 - CUP: E63C22001940006. This research has been partially supported by the project EXTRA-EYE - PRIN 2022 - CUP E53D23008280006 - Finanziato dall'Unione Europea - Next Generation EU.

% ---- Bibliography ----
%
% BibTeX users should specify bibliography style 'splncs04'.
% References will then be sorted and formatted in the correct style.
%
\bibliographystyle{splncs04}
\bibliography{main}

\clearpage
\author{Rosario Leonardi\inst{1}\orcidlink{0009-0001-8693-3826} \and Antonino Furnari\inst{1,2}\orcidlink{0000-0001-6911-0302} \and \\ Francesco Ragusa\inst{1,2}\orcidlink{0000-0002-6368-1910} \and Giovanni Maria Farinella\inst{1,2}\orcidlink{0000-0002-6034-0432}}

\authorrunning{R.~Leonardi et al.}
\institute{Department of Mathematics and Computer Science, University of Catania, Italy \and Next Vision s.r.l., Italy}

\title{Are Synthetic Data Useful for Egocentric Hand-Object Interaction Detection? (Supplementary Material)}
\titlerunning{Are Synthetic Data Useful for Egocentric HOI Detection? (Supp. Material)}
\maketitle
\appendix

\begin{figure}[ht]
    \centering
    \includegraphics[width=\linewidth]{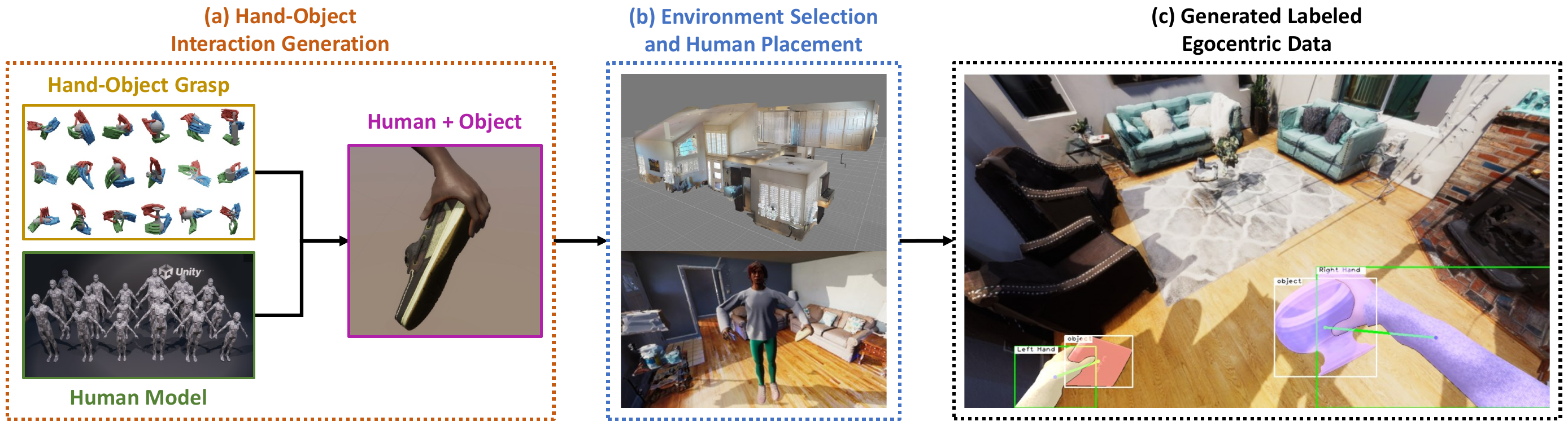}
    \caption{\textbf{The proposed data generation pipeline.} (a) An object-grasp pair is selected from DexGraspNet~\cite{wang2023dexgraspnet} 
    and integrated with a randomly generated human model. (b) The human + object model is placed in an environment randomly selected from the Habitat-Matterport 3D dataset~\cite{ramakrishnan2021habitat}. (c) Egocentric data of hand-object interactions is generated and automatically labeled. Labels include bounding boxes and segmentation masks of hands and interacted objects, contact-state, and hand-object relations.}
    \label{fig:simulator}
\end{figure}

\section{Additional Details on the proposed HOI-Synth Data Generation Pipeline and Simulator}\label{sec:add_simulator}
This section reports additional details on the proposed data generation pipeline and simulator (Figure~\ref{fig:simulator}). 
\begin{enumerate}
    \item We select a random hand-object grasp (Figure~\ref{fig:simulator}-a) from the \textit{DexGraspNet} dataset \cite{wang2023dexgraspnet}, a collection of 1.32 million robotic grasps generated with an accelerated differentiable force closure estimator. \textit{DexGraspNet} contains 5,355 untextured and convex-decomposed versions of 3D models from the ShapeNet~\cite{chang2015shapenet, savva2015semantically}, YCB~\cite{calli2017yale}, Big-BIRD~\cite{singh2014bigbird}, Grasp~\cite{kappler2015leveraging}, KIT~\cite{kasper2012kit} and Google's Scanned Objects~\cite{downs2022google}. We recover original textured objects from such datasets to obtain more realistic hand-object interactions. We also convert the hand pose included in \textit{DexGraspNet} from the ShadowHand format~\cite{shadowhand} to the \textit{SyntheticHumans} conventions.
    \item The hand-object grasp is fit to a randomly human model generated using \textit{SyntheticHumans} \cite{unity_synthetichumans_2022}, a package for the generation of rigged human 3D models. \textit{SyntheticHumans} offers different customization parameters for the generated human, such as ethnicity, clothing, age, height, weight, and sex.
    \item We choose a random indoor environment from the HM3D dataset~\cite{ramakrishnan2021habitat} (Figure~\ref{fig:simulator}-b). This dataset includes 1k reconstructions of real-world spaces like homes and stores. 
    \item We place the human-object model in the chosen environment at a random navmesh position, randomizing the pose of the hands and arms and placing a virtual camera at human eye level to capture the scene from the first-person point of view. 
\end{enumerate}

For each generated interaction, the \textit{Unity Perception} package~\cite{unity-perception2022} automatically annotates the bounding boxes and the segmentation masks of the hands and interacted objects. We extend Perception to also provide annotations for the hand contact state, and the hand-object relations as shown in Figure~\ref{fig:simulator}-c. 

\begin{figure*}[t]
    \centering
    \includegraphics[width=\linewidth]{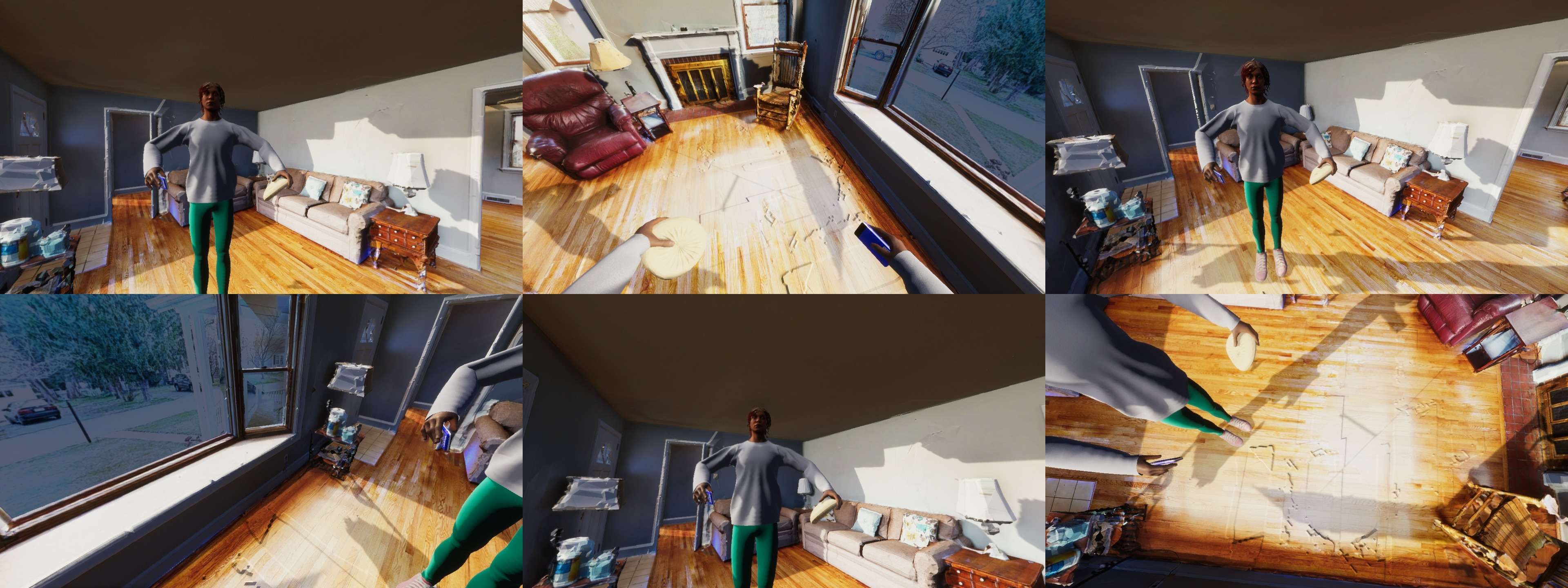}
    \caption{Examples of synthetic images captured from various points of view.}
    \label{fig:multi_point}
\end{figure*}

Additionally, our simulator allows the randomisation of different aspects of the virtual scene, such as light intensity, the color of light sources, object textures, etc. It supports the inclusion of visual effects like motion blur or noise to improve the realism and diversity of the scene. Furthermore, it is possible to increase the number and the position of virtual cameras, to increase the variability of the generated data and to consider even third-person points of view (see Figure~\ref{fig:multi_point}). 

\begin{figure*}[t]
    \centering
    \includegraphics[width=\linewidth]{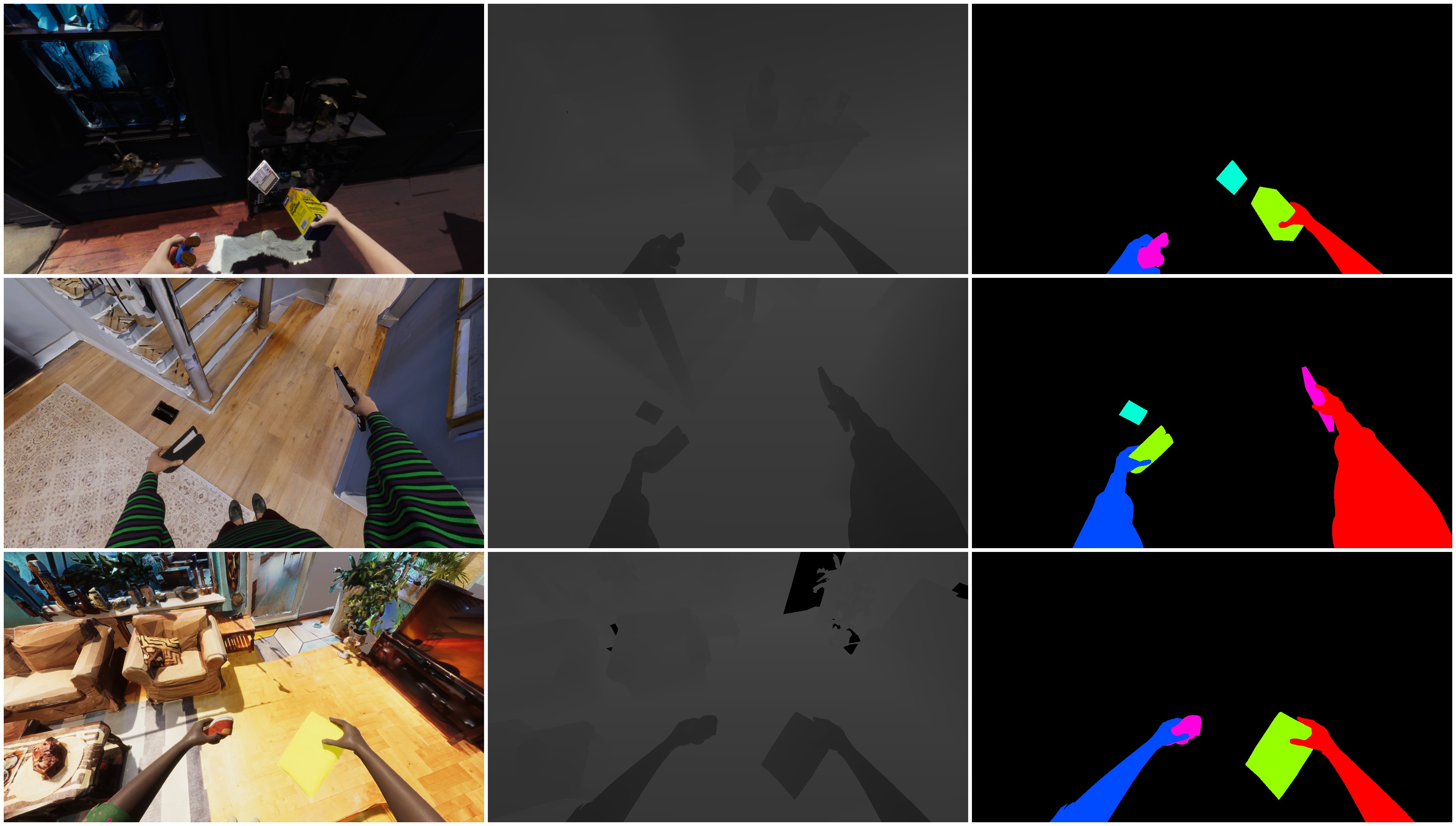}
    \caption{Examples of a synthetic image (left) with the corresponding depth map (center) and instance segmentation masks (right) generated with the proposed simulator.}
    \label{fig:multi_modalities}
\end{figure*}

Finally, the simulator enables the generation of synthetic video shots and additional labels such as depth maps (see Figure~\ref{fig:multi_modalities}), 3D bounding boxes, and 3D poses. While this work focused on the exploitation of the RGB modality, as in the standard HOS task, future works may also take advantage of the rich set of modalities and 3D annotations provided by our simulator.

See the attached video and images for examples of data generation.

\subsection{ENIGMA-51 synthetic data}
\begin{figure*}[t]
    \centering
    \includegraphics[width=\linewidth]{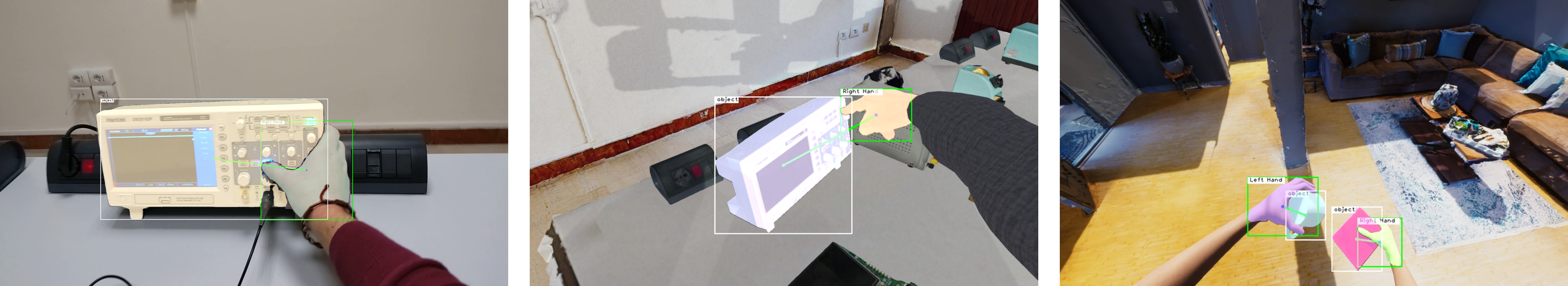}
    \caption{A ENIGMA-51 image (left), a synthetic in-domain image (center), and a synthetic out-domain image (right).}
    \label{fig:in-domain}
    
\end{figure*}

With the presented data generation pipeline, we complement the ENIGMA-51 dataset \cite{ragusa2024enigma} with two sets of synthetic images: an in-domain set and an out-domain set (Figure~\ref{fig:in-domain}). The in-domain set (Figure~\ref{fig:in-domain}-center) is generated using the 3D models of the environment and objects provided by the authors, thus obtaining synthetic images aligned to the real data. 
The out-domain set (Figure~\ref{fig:in-domain}-right) contains images of hand-object interactions in generic environments and with generic objects, akin to those generated to complement VISOR and EgoHOS.

\section{Implementation Details}
\begin{figure*}[!ht]
    \centering
    \includegraphics[width=\linewidth]{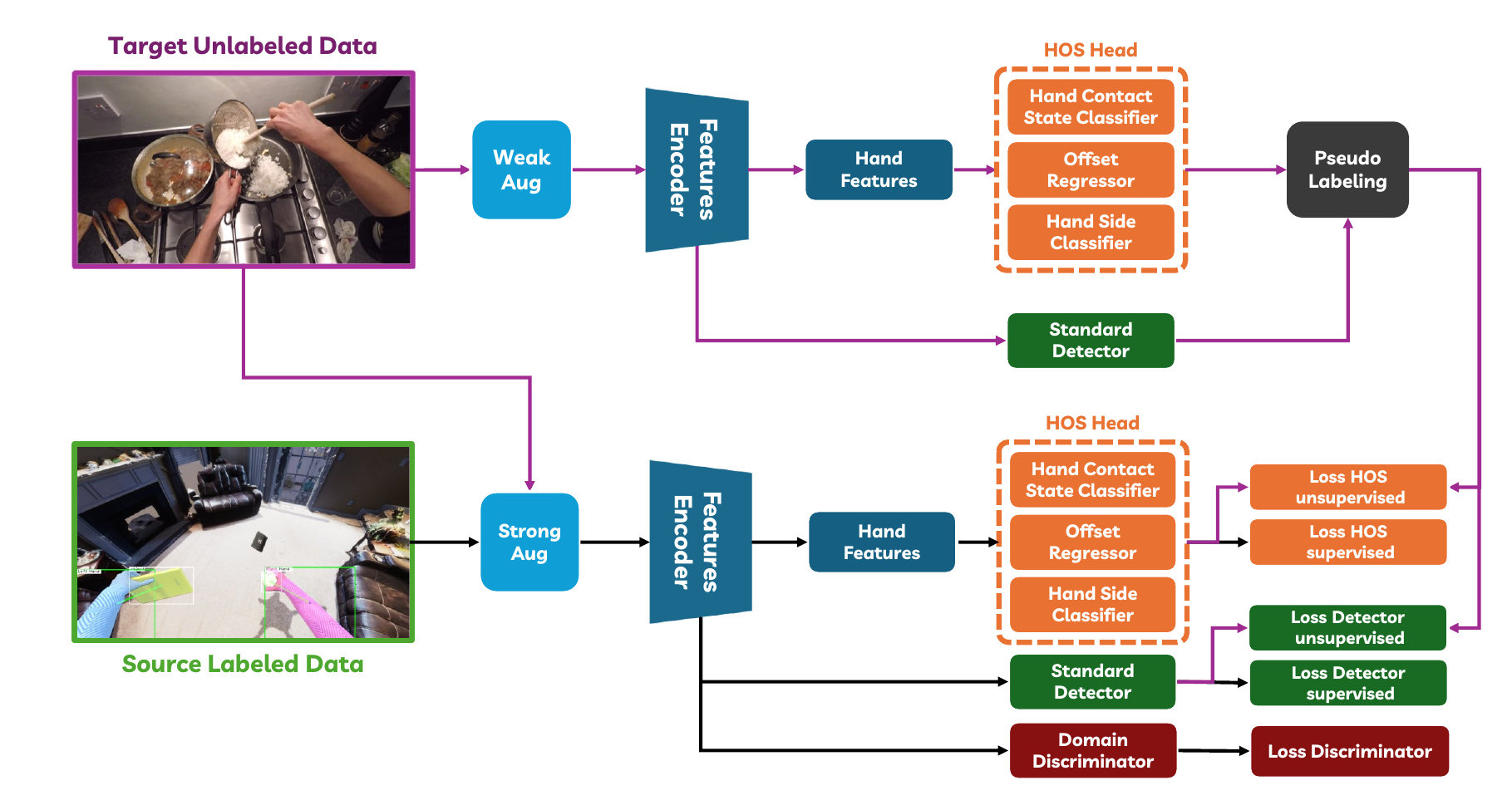}
    \caption{The architecture of the domain adaptation approach used in our analysis. The method is based on the Adaptive Teacher~\cite{li2022cross} framework, extended with HOS recognition modules~\cite{VISOR2022}. Refer to~\cite{li2022cross, VISOR2022} for more information on individual components and loss functions.}
    \label{fig:hos_at_arch}
\end{figure*}

\subsection{Domain Adaptation Methods}\label{sec:domain_adaptation}
We adapted four state-of-the-art semi-supervised domain adaptation methods for the considered HOS task. All considered domain adaptation methods are based on the teacher-student paradigm. Original methods have been adapted adding modules to estimate the hand side, contact state, offset vector, and segmentation masks as in~\cite{VISOR2022}. The considered methods are reported in the following.

\noindent\textbf{Mean Teacher (MT)~\cite{tarvainen2017mean}}. Mean Teacher is a semi-supervised learning technique based on the Teacher-Student paradigm. In this approach, both the student and teacher models process the same input applying different noise augmentation. During training, the weights are updated via gradient descent for the student's model, whereas the teacher weights are updated using an exponential moving average (EMA) of the student network's weights. We extend this method with additional modules for the HOS task.

\noindent\textbf{Mean Teacher + Adversarial Loss (MT+GRL)}. Following the idea of~\cite{li2022cross}, we combine the \textit{Mean Teacher} with \textit{Gradient Reverse Layer}~\cite{ganin2015unsupervised} (GRL) in order to align feature representations between source and target domains. We extend this method with additional modules for the HOS task.

\noindent\textbf{Unbiased Teacher (UT)~\cite{liu2021unbiased}}. In this work, the authors introduced a new training procedure composed of two steps. The first step is called \textit{Burn-In}, where a standard object detector is trained with labeled data, and the second step is called \textit{Teacher-Student Mutual Learning}, where the trained object detector is duplicated into a student and a frozen teacher model. The new unlabeled input images are passed to both models using different augmentation techniques. The teacher will produce pseudo-labels used to train the student model. We extend this method with additional modules for the HOS task.

\noindent\textbf{Adaptive Teacher (AT)~\cite{li2022cross}}. This method was originally designed to tackle cross-domain object detection. It combines the previously discussed techniques, i.e., GRL, Augmentation, and EMA. We extend this method with additional modules for the HOS task. The architecture of this approach is shown in Figure~\ref{fig:hos_at_arch}.

\subsection{Training details}\label{sec:training_details}
For consistency across the different approaches, we adopted the ResNet-101-FPN architecture as the backbone for our models. Tables~\ref{tab:training_visor}, \ref{tab:training_ego_hos}, and \ref{tab:training_enigma} show the hyperparameters used for training on EPIC-KITCHENS VISOR, EgoHos, and ENIGMA-51, respectively. These include learning rates, batch sizes, training iterations, batch sizes of the unlabeled images used in the domain adaptation approaches, and burn-up iterations. These values have been selected to optimize models' performance across the different datasets. Our models have been trained using four A30 NVIDIA GPUs and utilized the SGD optimizer. 

\begin{table*}[t]
    \centering
    \caption{Summary of hyperparameters used for training on EPIC-KITCHENS VISOR.}
    \resizebox{\linewidth}{!}{
        \begin{tabular}{c|lcccccc}
            \hline
            \textbf{\% Real Data} & \textbf{Approach} & \textbf{Batch Size} & \textbf{Iterations} & \textbf{LR} & \textbf{Batch Size Unlabeled} & \textbf{Burn-Up Iters} \\	
            \hline
            \multirow{2}{*}{0}    & Synthetic-Only    & 4                   & 80,000              & 0.01        & -                             & -                      \\
                                   & UDA               & 4                   & 100,000             & 0.0005      & 4                             & 20,000                 \\ \hline
                                                                                                                                                        
            \multirow{3}{*}{10}   & Real-Only         & 4                   & 20,000              & 0.01        & -                             & -                      \\
                                   & Synthetic+Real    & 4                   & 20,000              & 0.01        & -                             & -                      \\
                                   & SSDA              & 4                   & 100,000             & 0.0005      & 4                             & 20,000                 \\ \hline
                                                                                                                                                
            \multirow{3}{*}{25}   & Real-Only         & 4                   & 40,000              & 0.01        & -                             & -                      \\
                                   & Synthetic+Real    & 4                   & 40,000              & 0.01        & -                             & -                      \\
                                   & SSDA              & 4                   & 100,000             & 0.0005      & 4                             & 20,000                 \\ \hline
                                                                                                                                                
            \multirow{3}{*}{50}   & Real-Only         & 4                   & 50,000              & 0.01        & -                             & -                      \\
                                   & Synthetic+Real    & 4                   & 50,000              & 0.01        & -                             & -                      \\
                                   & SSDA              & 4                   & 100,000             & 0.0005      & 4                             & 20,000                 \\ \hline
                                                                                                                                                
            \multirow{3}{*}{100}  & Real-Only         & 4                   & 80,000              & 0.01        & -                             & -                      \\
                                   & Synthetic+Real    & 4                   & 80,000              & 0.01        & -                             & -                      \\
                                   & SSDA              & 4                   & 100,000             & 0.0005      & 4                             & 20,000                 \\ \hline
                                                                                                                                                
        \end{tabular}
    }
                                                                
    \label{tab:training_visor}
\end{table*}

\begin{table*}[t]
    \centering
    \caption{Summary of hyperparameters used for training on EgoHos.}

    \resizebox{\linewidth}{!}{
        \begin{tabular}{c|lcccccc}
            \hline
            \textbf{\% Real Data} & \textbf{Approach} & \textbf{Batch Size} & \textbf{Iterations} & \textbf{LR} & \textbf{Batch Size Unlabeled} & \textbf{Burn-Up Iters} \\	
            \hline
            \multirow{2}{*}{0}    & Synthetic-Only    & 4                   & 80,000              & 0.01        & -                             & -                      \\
                                   & UDA               & 4                   & 80,000              & 0.0005      & 4                             & 20,000                 \\ \hline
                                                                                                                                                        
            \multirow{3}{*}{10}   & Real-Only         & 4                   & 20,000              & 0.01        & -                             & -                      \\
                                   & Synthetic+Real    & 4                   & 20,000              & 0.01        & -                             & -                      \\
                                   & SSDA              & 4                   & 80,000              & 0.0005      & 4                             & 20,000                 \\ \hline
                                                                                                                                                
            \multirow{3}{*}{25}   & Real-Only         & 4                   & 40,000              & 0.01        & -                             & -                      \\
                                   & Synthetic+Real    & 4                   & 40,000              & 0.01        & -                             & -                      \\
                                   & SSDA              & 4                   & 80,000              & 0.0005      & 4                             & 20,000                 \\ \hline
                                                                                                                                                
            \multirow{3}{*}{50}   & Real-Only         & 4                   & 50,000              & 0.01        & -                             & -                      \\
                                   & Synthetic+Real    & 4                   & 50,000              & 0.01        & -                             & -                      \\
                                   & SSDA              & 4                   & 80,000              & 0.0005      & 4                             & 20,000                 \\ \hline
                                                                                                                                                
            \multirow{3}{*}{100}  & Real-Only         & 4                   & 80,000              & 0.01        & -                             & -                      \\
                                   & Synthetic+Real    & 4                   & 80,000              & 0.01        & -                             & -                      \\
                                   & SSDA              & 4                   & 80,000              & 0.0005      & 4                             & 20,000                 \\ \hline
                                                                                                                                                
        \end{tabular}
    }
                                                                
    \label{tab:training_ego_hos}
\end{table*}

\begin{table*}[t]
    \centering
    \caption{Summary of hyperparameters used for training on ENIGMA-51.}
    \resizebox{\linewidth}{!}{
        \begin{tabular}{c|lcccccc}
            \hline
            \textbf{\% Real Data} & \textbf{Approach} & \textbf{In-domain} & \textbf{Batch Size} & \textbf{Iterations} & \textbf{LR} & \textbf{Batch Size Unlabeled} & \textbf{Burn-Up Iters} \\	
            \hline
            \multirow{4}{*}{0}    & Synthetic-Only    &                    & 4                   & 50,000              & 0.01        & -                             & -                      \\
                                  & Synthetic-Only    & \checkmark         & 4                   & 50,000              & 0.01        & -                             & -                      \\
                                  & UDA               &                    & 4                   & 50,000              & 0.0005      & 4                             & 20,000                 \\
                                  & UDA               & \checkmark         & 4                   & 50,000              & 0.0005      & 4                             & 20,000                 \\ \hline
                                                                                                                                                                                                                                            
            \multirow{5}{*}{10}   & Real-Only         &                    & 4                   & 20,000              & 0.01        & -                             & -                      \\
                                  & Synthetic+Real    &                    & 4                   & 20,000              & 0.01        & -                             & -                      \\
                                  & Synthetic+Real    & \checkmark         & 4                   & 20,000              & 0.01        & -                             & -                      \\
                                  & SSDA              &                    & 4                   & 50,000              & 0.0005      & 4                             & 20,000                 \\
                                  & SSDA              & \checkmark         & 4                   & 50,000              & 0.0005      & 4                             & 20,000                 \\ \hline
                                                                                                                                                                                                                                    
            \multirow{5}{*}{25}   & Real-Only         &                    & 4                   & 30,000              & 0.01        & -                             & -                      \\
                                  & Synthetic+Real    &                    & 4                   & 30,000              & 0.01        & -                             & -                      \\
                                  & Synthetic+Real    & \checkmark         & 4                   & 30,000              & 0.01        & -                             & -                      \\
                                  & SSDA              &                    & 4                   & 50,000              & 0.0005      & 4                             & 20,000                 \\
                                  & SSDA              & \checkmark         & 4                   & 50,000              & 0.0005      & 4                             & 20,000                 \\ \hline
                                                                                                                                                                                                                                    
            \multirow{5}{*}{50}   & Real-Only         &                    & 4                   & 40,000              & 0.01        & -                             & -                      \\
                                  & Synthetic+Real    &                    & 4                   & 40,000              & 0.01        & -                             & -                      \\
                                  & Synthetic+Real    & \checkmark         & 4                   & 40,000              & 0.01        & -                             & -                      \\
                                  & SSDA              &                    & 4                   & 50,000              & 0.0005      & 4                             & 20,000                 \\
                                  & SSDA              & \checkmark         & 4                   & 50,000              & 0.0005      & 4                             & 20,000                 \\ \hline
                                                                                                                                                                                                                                    
            \multirow{5}{*}{100}  & Real-Only         &                    & 4                   & 50,000              & 0.01        & -                             & -                      \\
                                  & Synthetic+Real    &                    & 4                   & 50,000              & 0.01        & -                             & -                      \\
                                  & Synthetic+Real    & \checkmark         & 4                   & 50,000              & 0.01        & -                             & -                      \\
                                  & SSDA              &                    & 4                   & 50,000              & 0.0005      & 4                             & 20,000                 \\
                                  & SSDA              & \checkmark         & 4                   & 50,000              & 0.0005      & 4                             & 20,000                 \\ \hline
                                                                                                                                                                                                                                    
        \end{tabular}
    }
                                                                                            
    \label{tab:training_enigma}
\end{table*}

\section{Additional Experiments}
\subsection{Explicit contact and side metrics}
We further computed F1 scores for hand contact and side predictions, regardless of whether the hand segmentation is accurate. This approach ensures that our evaluation focuses on the core tasks without being influenced by the quality of the hand segmentation. Results in Table~\ref{tab:f1_score} shows how integrating synthetic data consistently enhances performance, demonstrating the benefits of simulating interactions beyond merely improving semantic segmentation. For the VISOR dataset, the Contact F1 score increases from $79.46\%$ to $81.65\%$, and the Side F1 score improves from $98.92\%$ to $99.06\%$. Similar improvements are observed for the EgoHOS dataset, with the Contact F1 score rising from $78.78\%$ to $79.25\%$ and the Side F1 score from $98.33\%$ to $98.87\%$. The ENIGMA-51 dataset shows an increase in Contact F1 score from $77.28\%$ to $77.83\%$ and in Side F1 score from $98.80\%$ to $99.11\%$. These consistent gains highlight the robustness and effectiveness of using synthetic data in enhancing model performance for contact and side prediction.
\begin{table*}[t]
    \centering
    \caption{Comparison of FSDA vs Real-Only for F1 Scores of Contact and Side.}
    \resizebox{0.7\linewidth}{!}{
    		\begin{tabular}{lccc}
    			\hline
    			\textbf{Dataset} & \textbf{Approach} & \textbf{Contact F1} & \textbf{Side F1} \\
    			\hline               
    			VISOR            & Real-Only         &  79.46           &   98.92       \\
    			VISOR            & FSDA              &  \textbf{81.65}           &  \textbf{99.06}        \\ 
    			\hline 
    			EgoHOS          & Real-Only         &  78.78           & 98.33         \\ 
    			EgoHOS          & FSDA              &  \textbf{79.25}           & \textbf{98.87}          \\ 
    			\hline
    			ENIGMA-51        & Real-Only         &  77.28            & 98.80              \\ 
    			ENIGMA-51        & FSDA              &  \textbf{77.83}             & \textbf{99.11}              \\     
    						            
    			\hline
    		\end{tabular}
    	}
         \label{tab:f1_score}
\end{table*}

\subsection{Simulation vs classic synthetic augmentations techniques}
\begin{figure*}[!ht]
    \centering
    \includegraphics[width=\linewidth]{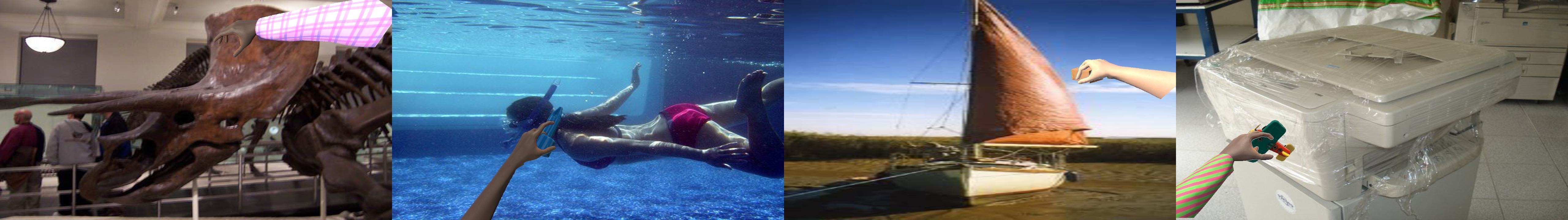}
    \caption{Simple HOI generation approach: Object models and hands are randomly projected onto ImageNet backgrounds.}
    \label{fig:sim_vs_aug}
\end{figure*}

\begin{table*}[t]
     \centering
     \caption{UDA Results on VISOR: Comparison of simulation-based image generation and classic synthetic augmentations.}

    \resizebox{\linewidth}{!}{
        \begin{tabular}{l|c|cccc}
            \hline
            \textbf{Generation approach} & \textbf{Overall} & H              & H+C            & O             & Contact F1     \\
            \hline               
            Synthetic augmentation       & 19.62            & 62.40          & 11.91          & 3.60          & 8.78           \\
            Simulation (Ours)            & \textbf{30.54}   & \textbf{79.05} & \textbf{31.51} & \textbf{7.33} & \textbf{22.75} \\ 
                                                                        
            \hline
        \end{tabular}
    }
    \label{tab:sim_vs_aug}

\end{table*}

Table~\ref{tab:sim_vs_aug} shows \textit{Synthetic} $\to$ \textit{VISOR} UDA results when using our images generated in simulation vs classic synthetic augmentations techniques on a subset of 3,500 images. Synthetic augmentation images are obtained by randomly projecting object models and hands on random background images from ImageNet (see Figure~\ref{fig:sim_vs_aug}). This experiment aims to demonstrate that images generated through simulation, which accurately model hand-object interactions, lead to superior performance. Results show clear advantages of simulated images, especially in \textit{H+C} and \textit{Contact F1} metrics. Specifically, the simulation approach outperforms the simple HOI generation approach across all the considered metrics, demonstrating the importance of realistic interaction modeling in improving the performance of models.

\subsection{Object Overlap Analysis}
The analysis of overlaps in object classes between the generated data and each dataset provides insights into the alignment and relevance of the datasets for in-domain studies. Specifically, the overlap percentages are as follows: 46.6\% (\textit{VISOR}), 29.1\% (\textit{EgoHos}), 0.4\% (\textit{ENIGMA-51}). These percentages indicate the extent to which the object classes present in the generated data are also found in each of the existing datasets. The results obtained in each dataset, especially in the UDA setting, indicate that the effectiveness of synthetic data varies significantly based on the overlap of object classes, suggesting the importance of considering in-domain studies, particularly for datasets like \textit{ENIGMA-51} with minimal overlap.

\subsection{Targeting Field Of View on ENIGMA-51}
The table~\ref{tab:fov_enigma} shows results when aligning the Field Of View (FOV) of out-domain synthetic data to that of \textit{ENIGMA-51}. This alignment aims to investigate whether adjusting the FOV can enhance the performance of methods. The results indicate that FOV can indeed have a slight impact on method performance. However, it is evident from the obtained results that, while targeting FOV is beneficial, the performance gap between out-domain synthetic data and in-domain data remains significant. This highlights the critical importance of using in-domain synthetic data to achieve optimal performance within a specific domain.
\begin{table*}[!ht]
    \centering
    \caption{Performance comparison when aligning the FOV of out-domain synthetic data to that of \textit{ENIGMA-51}. }
    \resizebox{\linewidth}{!}{
        \begin{tabular}{lcc|c|ccc}
            \hline
            \textbf{Approach} & \textbf{Target FOV} &\textbf{In-domain} & \textbf{Overall} &  H              & H+C            & O              \\
            \hline               
            Synthetic-Only    &  &                            & 00.21 & 01.07 & 00.03 & 00.99           \\
            Synthetic-Only    & \checkmark &                  & 05.67             & 15.78          & 02.66           & 02.31           \\
            Synthetic-Only    &\checkmark& \checkmark         & \textbf{12.85}   & \textbf{56.05} & \textbf{15.24} & \textbf{4.79}  \\ \hline
        \end{tabular}
    }
    \label{tab:fov_enigma}
\end{table*}

\subsection{Generalization of Findings with Different Backbone Networks} 
To evaluate the robustness and generality of our approach, we replicated experiments using a \textit{ResNet50} backbone instead of the originally employed \textit{ResNet101}. Our experiments in the Unsupervised Domain Adaptation (UDA) setting yielded an Overall Average Precision of $27.68\%$ with the \textit{ResNet50} backbone, which is a significant improvement of $+17.80\%$ compared to the baseline. In comparison, the \textit{ResNet101} backbone achieved an AP of $33.33\%$, showing an improvement of $+23.45\%$ relative to the baseline. These results suggest that our findings are not only specific to a particular model configuration but also generalizable across different network architectures, reinforcing the effectiveness of using synthetic data.

\subsection{Validation of Synthetic Data Volume} 
Figure~3 from the main paper indicates that the volume of synthetic data is adequate, as evidenced by the approaching plateau in performance metrics. To further confirm this, we conducted an additional experiment on \textit{VISOR} in which we trained our UDA approach with 80,000 synthetic images. The results show that we achieved an Overall Average Precision of $33.19\%$, which closely aligns with previous findings of $33.33\%$. This consistency confirms the sufficiency of the synthetic data volume for effective model training.

\subsection{Synthetic vs real data for UDA} 
Table~\ref{tab:synth_vs_out_domain_real} provides a comparison of Unsupervised Domain Adaptation (UDA) on \textit{EgoHOS} when using either our synthetic data or \textit{EPIC-KITCHENS VISOR} data as the source domain. To ensure a fair evaluation, we have excluded \textit{EgoHOS} images that overlap with \textit{EPIC-KITCHENS} to avoid duplicity with \textit{VISOR}. Results show clear advantages of using synthetic data, especially on \textit{Contact} and \textit{Overall} metrics, while results are comparable on \textit{O} and \textit{H}. These findings underscore the critical role of high-quality examples of hand-object interactions included in our synthetic data and reinforce the efficacy of leveraging synthetic data for domain adaptation in egocentric vision applications.

\begin{table*}[t]
    \centering
    \caption{Comparison of Unsupervised Domain Adaptation (UDA) on \textit{EgoHOS} using either synthetic data or \textit{EPIC-KITCHENS VISOR} data as the source domain. \textit{EgoHOS} images sampled from \textit{EPIC-KITCHENS} are excluded to prevent overlap with \textit{VISOR}.}
    \resizebox{0.9\linewidth}{!}{
		\begin{tabular}{lc|c|ccc}
			\hline
			\textbf{Training Data} & \textbf{Test Data} & \textbf{Overall} & H & H+C & O \\
			\hline               
			Synthetic Data                          & EgoHOS             & \textbf{29.38}            & 69.65      & \textbf{21.01}        & \textbf{10.79}      \\
			VISOR                                   & EgoHOS             & 24.39            & \textbf{71.97}      & 13.43        & 10.59      \\  \hline
		\end{tabular}
    }
    \label{tab:synth_vs_out_domain_real}
 \end{table*}

\subsection{Additional Qualitative Experiments}\label{sec:add_qualitative}
\begin{figure*}[t]
    \centering
    \includegraphics[width=\linewidth]{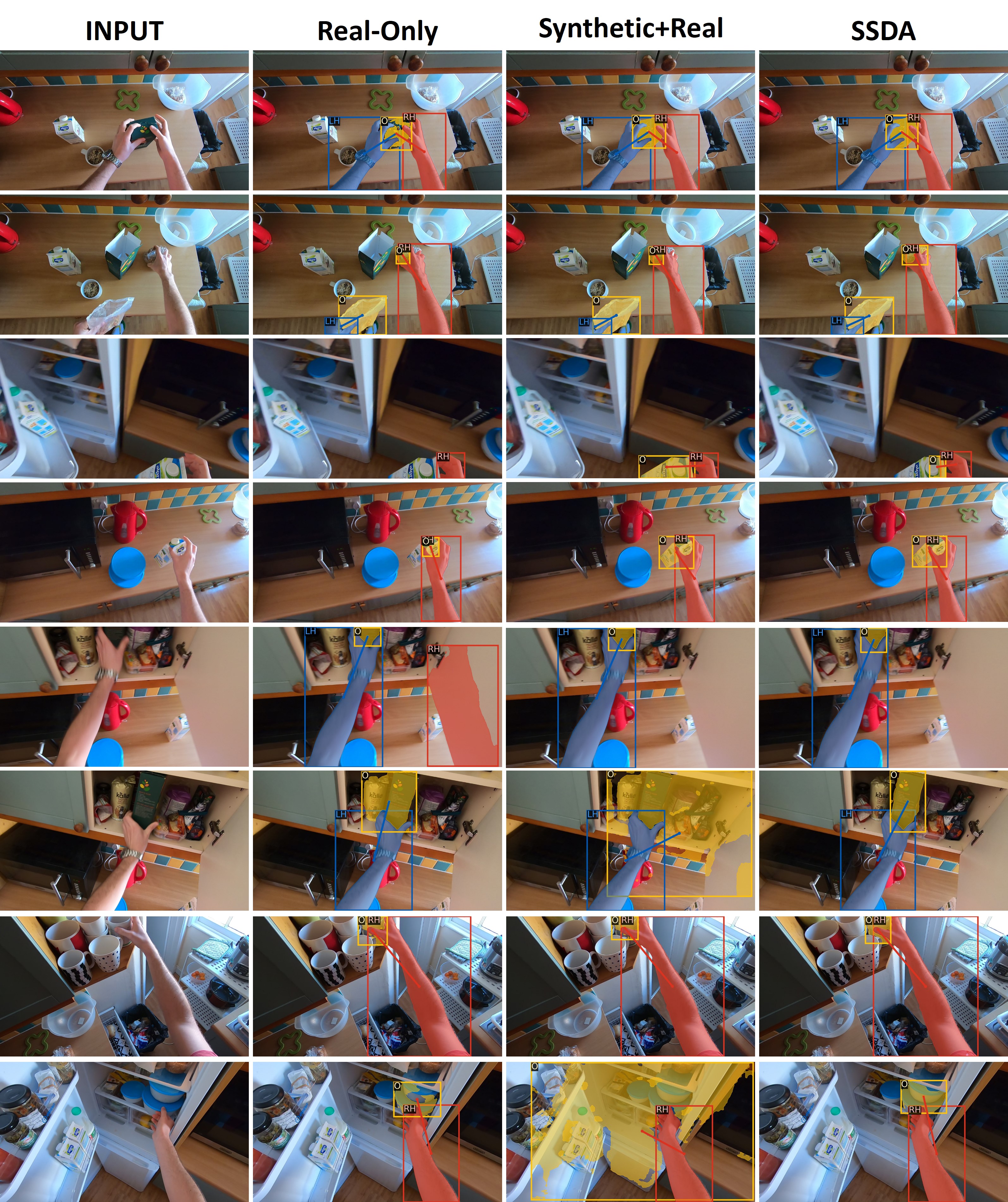}
    \caption{Additional qualitative examples on VISOR.}
    \label{fig:qualitative_examples_visor}
\end{figure*}

\begin{figure*}[t]
    \centering
    \includegraphics[width=\linewidth]{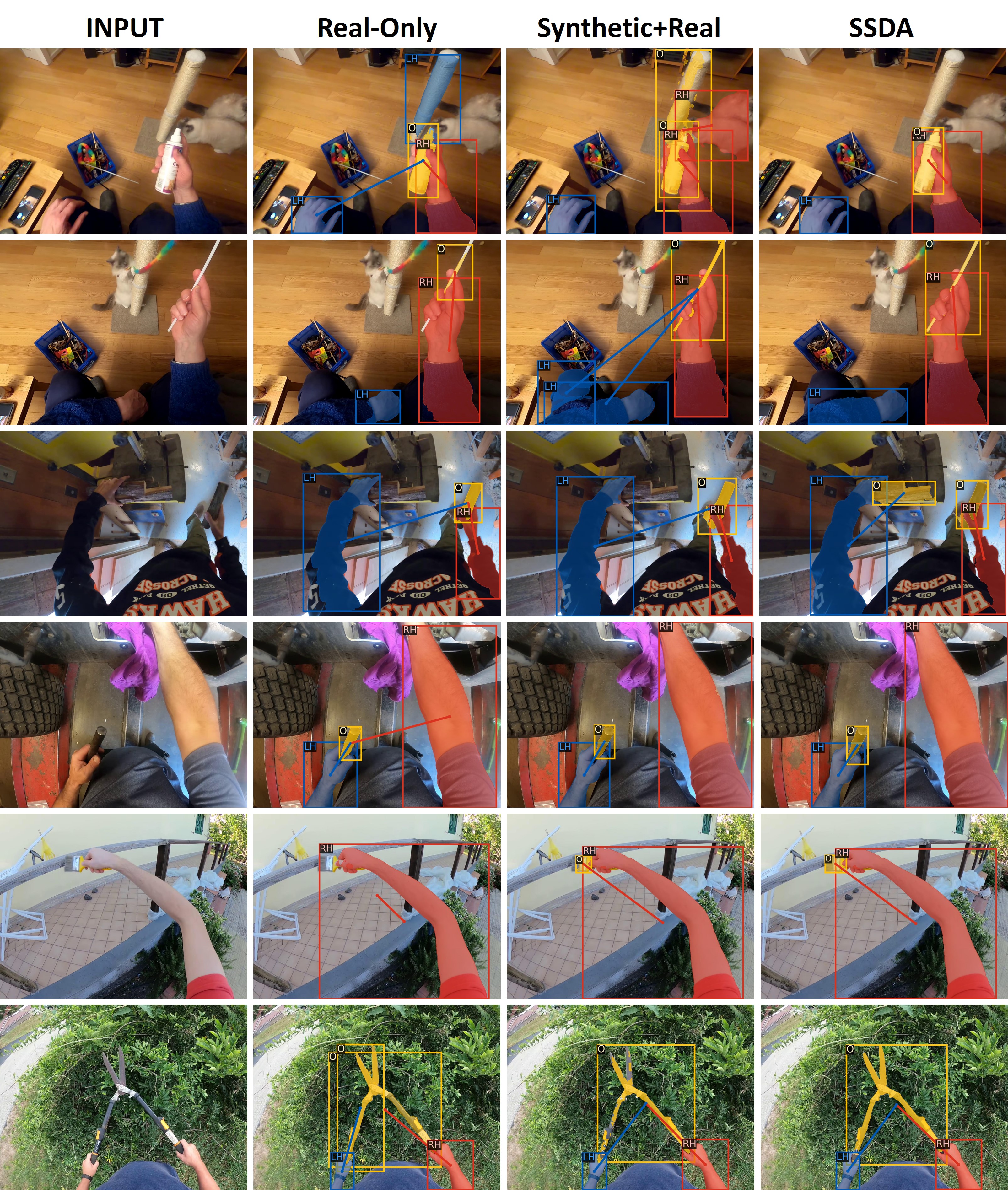}
    \caption{Additional qualitative examples on EgoHos.}
    \label{fig:qualitative_examples_egohos}
\end{figure*}

\begin{figure*}[t]
    \centering
    \includegraphics[width=\linewidth]{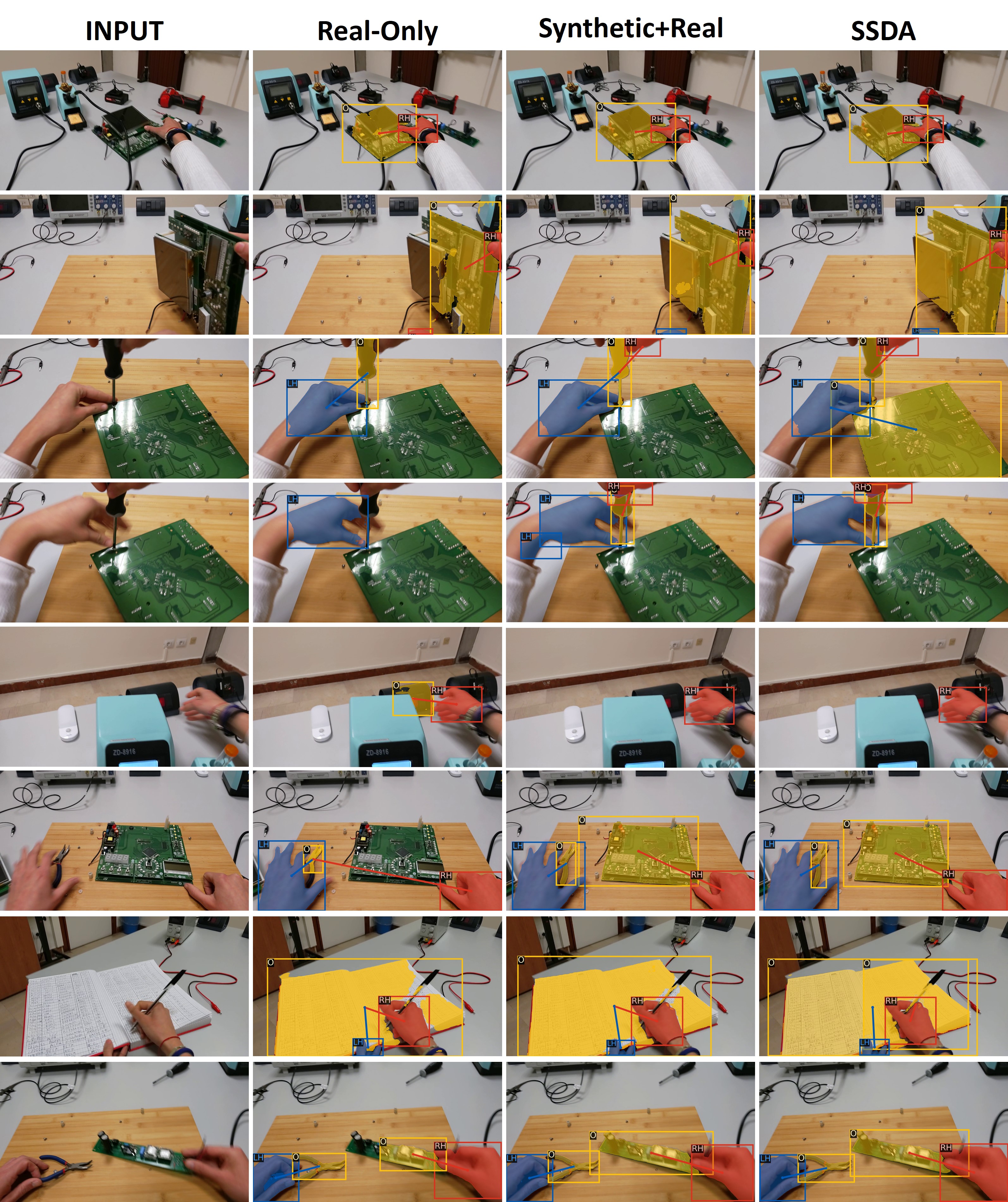}
    \caption{Additional qualitative examples on Enigma-51.}
    \label{fig:qualitative_examples_enigma}
\end{figure*}

Figures~\ref{fig:qualitative_examples_visor}, \ref{fig:qualitative_examples_egohos}, and \ref{fig:qualitative_examples_enigma} show qualitative examples obtained by the different approaches on EPIC-KITCHENS VISOR, EgoHos, and ENIGMA-51 datasets respectively. In particular, looking at the VISOR examples (Figure~\ref{fig:qualitative_examples_visor}), we can notice that the \textit{SSDA} approach (fourth column) obtains better object segmentation than \textit{Real-Only} (second column) and \textit{Synthetic+Real} (third column) in almost all cases except for the third line in which the \textit{Synthetic+Real} achieves more precise segmentation. Note that for the example in the third row, the \textit{Real-Only} method fails to detect the interacted object. Similarly, in the EgoHos examples (Figure~\ref{fig:qualitative_examples_egohos}), \textit{SSDA} constantly improves the segmentation and reduces false positives (see first, second, third, and last rows). Again, the \textit{Real-Only} approach fails to detect the interacted object (paintbrush) in the fifth row. Finally, considering the ENIGMA-51 dataset (Figure~\ref{fig:qualitative_examples_enigma}), the \textit{SSDA} approach significantly improves the segmentation in all the examples and reduces the false positive (see third, fourth, fifth, sixth, and seventh rows) compared to the \textit{Real-Only} approach.

\end{document}